\documentclass[final]{clv2025}

\jvol{vv}
\jnum{nn}
\jyear{2025}

\dochead{Long Paper} 

\pageonefooter{Action editor: \{action editor name\}. Submission received: DD Month YYYY; revised version received: DD Month YYYY; accepted for publication: DD Month YYYY.}

\usepackage{booktabs}
\usepackage{amsmath}

\usepackage{graphicx}
\usepackage{multirow}
\usepackage{adjustbox}
\usepackage{url}
\usepackage{amsmath}
\usepackage{float}
\usepackage{arabtex}
\usepackage{utf8}
\makeatletter
\protected\def\begin#1{%
  \UseHook{env/#1/before}%
  \@ifundefined{#1}%
    {\def\reserved@a{\@latex@error{Environment #1 undefined}\@eha}}%
    {\def\reserved@a{\def\@currenvir{#1}%
        \edef\@currenvline{\on@line}%
        \@execute@begin@hook{#1}%
        \csname #1\endcsname}}%
  \@ignorefalse
  \begingroup
  \let\end\a@l@end 
  \@endpefalse\reserved@a}
\makeatother

\setcode{utf8}

\runningtitle{Character Iconicity vs. Arbitrariness}
\runningauthor{Alomari, Ahmad, and Al-shaibani}

\begin{document}

\title{Character Iconicity vs. Arbitrariness: \\An Arabic NLP Perspective}

\author{Dorieh Alomari\thanks{Corresponding author}$^{,1,2}$, Irfan Ahmad$^{1,2}$, and Maged S. Al-shaibani$^{2}$}

\affilblock{
    \affil{King Fahd University of Petroleum and Minerals, Saudi Arabia\\\quad \email{g202213300@kfupm.edu.sa}, \email{irfan.ahmad@kfupm.edu.sa}}
    \affil{SDAIA--KFUPM JRC for AI, Saudi Arabia\\\quad \email{majed.alshaibani@kfupm.edu.sa}}
}

\maketitle

\begin{abstract}
Arabic script uses 28 letters, many of which share a common base shape (rasm) and are distinguished only by dot placement. Because early Arabic manuscripts were written without dots yet remained interpretable, dot removal offers a natural test of whether these visual distinctions are functionally necessary. 
Prior work has shown that dotless Arabic can remain readable and effective for natural language processing (NLP), but it remains unclear whether this success depends on preserving the original rasm groupings or whether arbitrary but consistent remappings to the same reduced rasm set can achieve comparable performance. We address this question by comparing standard dotted and dotless Arabic with arbitrary character remappings constrained to the same 19 undotted rasms. We generated 2,000 random remappings under word- and character-level tokenization and selected four representative mappings with the highest and lowest entropy values. These representations were evaluated across language modeling, text classification, sequence labeling, machine translation, and restoration to the original script. The results show that neither preserving original character distinctions nor retaining traditional rasm-based groupings is necessary for strong NLP performance. Random remappings achieve competitive performance while reducing vocabulary size, out-of-vocabulary (OOV) rates, model size, and training cost. These findings suggest that, from an NLP perspective, Arabic character form-function relationships are largely arbitrary: models rely more on stable distributional structure than on the visual iconicity of letter forms.\footnote{Our code and data are available at: (GitHub link will be released upon acceptance).}
\end{abstract}

\section{Introduction}

Linguistic theory has traditionally emphasized the arbitrariness of orthographic form, arguing that the relationship between written characters and meaning is largely arbitrary \citep{hockett1960origin, saussure1983course}. However, more recent research challenges a strictly arbitrary view, proposing that iconicity---systematic correspondences between aspects of form and meaning---coexists with arbitrariness in shaping language structure \citep{dingemanse2015arbitrariness}. Such correspondences may link perceptual experience with linguistic representation through structural or multimodal patterns \citep{perniss2014bridge, hodge2022iconicity}. This debate raises a key question for writing systems: do the visual properties of characters have functional significance, or are they computationally arbitrary?

These theoretical developments raise an important question for computational modeling: when systematic form-meaning patterns appear in data, are they rooted in meaningful structural properties of symbols, or do models simply exploit statistical regularities regardless of underlying design? Arabic script provides a compelling test case. Because many letters share identical base shapes and differ only by dots, the script naturally encodes visual groupings. If these groupings reflect deeper structural or cognitive organization, preserving them should influence NLP performance. Conversely, if performance remains stable under arbitrary remappings, this would support the arbitrariness hypothesis from a computational perspective.

Prior work (such as \citet{al2023dotless}) has shown that dotless Arabic is computationally viable: many NLP tasks can be performed competitively even when Arabic letters that share the same rasm are collapsed into a single undotted form. However, this leaves open a deeper question: does dotless Arabic work because the original rasm groupings preserve meaningful visual or orthographic structure, or because NLP models can adapt to alternative reduced symbol systems as long as the mapping is stable? This paper tests these alternatives directly. By comparing standard dotless Arabic with rasm-constrained random remappings, we show that preserving the historical rasm-based groupings is not necessary for strong NLP performance.

\begin{figure}[ht]
    \centering
    \includegraphics[width=0.45\linewidth]{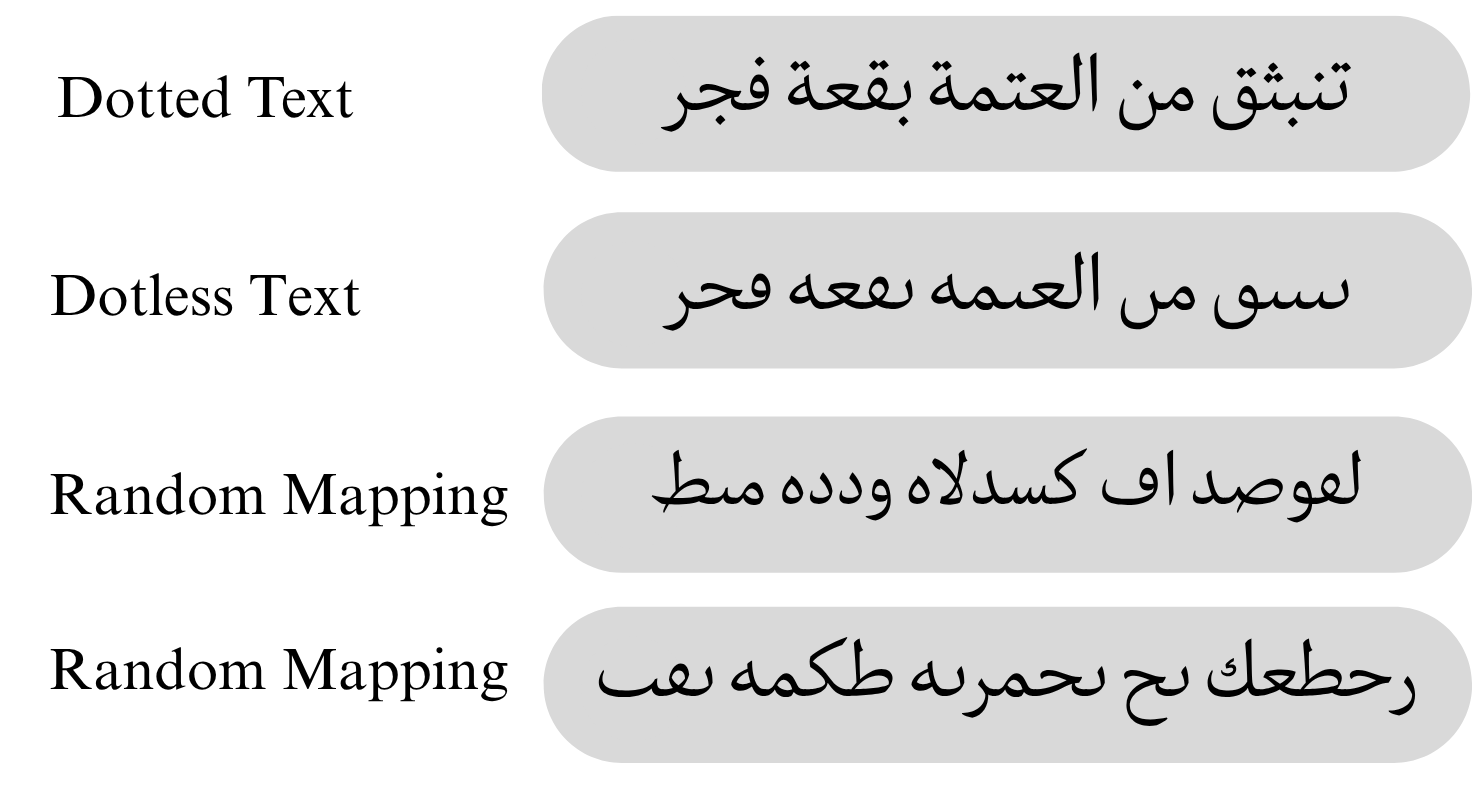}
    \caption{The dotless version demonstrates the reduction of letters to the base rasms, while the random mappings show how letters can be reassigned to these rasms arbitrarily.}
    \label{fig:rm-sample}
\end{figure}

A sample sentence in standard dotted form, dotless form, and two alternative random mappings is shown in Figure~\ref{fig:rm-sample}. Our first objective is to test whether shared rasm groupings maintain performance due to underlying orthographic similarities, or whether models perform equally well under arbitrary remappings. Our second objective is practical: we evaluate whether particular remapping strategies reduce vocabulary size or improve computational efficiency compared to the standard dotless reduced character set, without compromising accuracy, thereby yielding improved trade-offs between representational simplicity and overall performance.

This paper makes the following contributions:
\begin{itemize}
    \item We introduce a controlled comparison between standard dotless Arabic and rasm-constrained random character remappings to test whether the success of dotless Arabic in NLP reflects the visual iconicity of historically motivated rasm groupings, or whether models can rely on arbitrary but stable reduced-script mappings.
    
    \item We evaluate standard dotted Arabic, dotless Arabic, and rasm-constrained random remappings across language modeling, text classification, sequence labeling, machine translation, and text restoration, showing that Arabic NLP models can often maintain competitive performance even when traditional visual letter groupings are disrupted.

    \item We demonstrate that reduced-script representations can substantially reduce vocabulary size and model size, and in several settings reduce training time, while learned restoration models make it feasible to recover fully specified Arabic text after processing simplified or remapped input.
\end{itemize}


The remainder of the paper is organized as follows: Section \ref{sec:Related} reviews related work, Section \ref{sec:Methodology} details the proposed methodology, Section \ref{sec:Experiments} reports experiments and results, and Section \ref{sec:Conclusion} concludes with key findings and future directions.



\section{Related Work}
\label{sec:Related}

Growing empirical evidence challenges the view that written forms are purely arbitrary. Large-scale analyses show that letter shapes reflect perceptual constraints and resemble structural patterns in natural scenes \citep{changizi2006structures}. Studies of graphic complexity reveal systematic variation across writing systems and predictable historical compression of letter forms \citep{miton2021complexity, kelly2021predictable}. Quantifiable visual features such as angularity align with human perception \citep{porto2025glyph}, and experimental as well as computational work demonstrates that orthographic form influences processing and encodes structured similarity patterns \citep{roman2024siamese, sidhu2025orthographic, taylor2025optimal}. 

Iconicity has also been examined through sound-shape correspondences, with cross modal experiments confirming robust sound-shape mappings across cultures \citep{dingemanse2016sound, turoman2017glyph, cwick2021bouba}, and evidence that graphemic properties, such as font angularity, affect perceptual processing \citep{decarolis2018implicit}. Advances in Artificial Intelligence (AI) further show that neural models can recover phonetic or semantic patterns from orthographic input \citep{hartmann2019predicting, li2022zero, loakman2024multimodal}, raising the question of whether systematic correspondences stem from perceptual design constraints or distributional regularities learnable by models.

Research on alternative Arabic text representations has primarily focused on dot removal. Pretrained models maintain competitive performance on undotted Arabic with appropriate tokenization \citep{rom2021supporting}, and neural methods can reliably restore dots \citep{alhathloul2022automatic}. Notably, \citet{al2023dotless} demonstrated that dotless Arabic reduces vocabulary size and OOV rates with minimal entropy loss while preserving performance, and \citet{al2023consonant} showed similar robustness for consonant-based reductions in English.


Despite these advances, prior research has focused on structured simplifications grounded in traditional rasm groupings. The broader question remains unexplored: does the success of dotless text depend on preserving visually related character clusters, or can entirely random character mappings achieve comparable performance? Addressing this gap enables a direct test of iconicity versus arbitrariness in Arabic character representations.

Building upon \citet{al2023dotless}, our work extends the investigation by systematically comparing random character remappings. By evaluating their impact on information content and NLP performance, we aim to determine whether Arabic character form-function relationships reflect meaningful orthographic similarities or primarily statistical learnability.

\section{Methodology}
\label{sec:Methodology}

\begin{figure}[ht]
    \centering
    \includegraphics[width=0.4\linewidth]{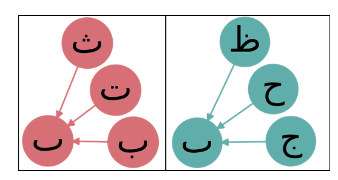}
    \caption{A sample illustrating the mapping of three letters to a standard rasm form (left), alongside a set of three randomly assigned characters mapped to the same rasm (right).}
    \label{fig:undotted-example}
\end{figure}

To investigate whether the relationship between characters and their shapes matters, or whether arbitrary and random mappings are equally effective, we analyze the effects of randomly remapping Arabic characters and evaluate their impact across several NLP tasks. 
To implement this approach, we use the 19 rasms of the standard dotless Arabic mapping as the fixed output symbol set. Each random mapping preserves the original rasm group sizes, but randomly changes which letters are assigned to each rasm. For example, in the standard dotless mapping, \<ب>, \<ت>, and \<ث> all map to \<ٮ>; in our random mappings, three randomly selected letters are assigned to \<ٮ>, as illustrated in Figure~\ref{fig:undotted-example}.



Using word-level and character-level tokenizers, we generated 2,000 random mappings for each and selected the mappings with maximum and minimum entropy. The chosen mappings are evaluated on language modeling, text classification, sequence labeling, and machine translation. A restoration model is also developed to recover the original dotted text when required. Multiple tokenization strategies were incorporated into the experiments to evaluate the alternative mappings across different tokenization schemes, including word tokenizer, character tokenizer, Farasa Morphological tokenizer \citep{abdelali2016farasa}, disjoint-letters tokenizer \citep{alyafeai2023evaluating}, and SentencePiece Byte Pair Encoding (BPE) \citep{kudo2018sentencepiece}. 

\begin{figure}[ht]
    \centering
    \includegraphics[width=0.85\linewidth]{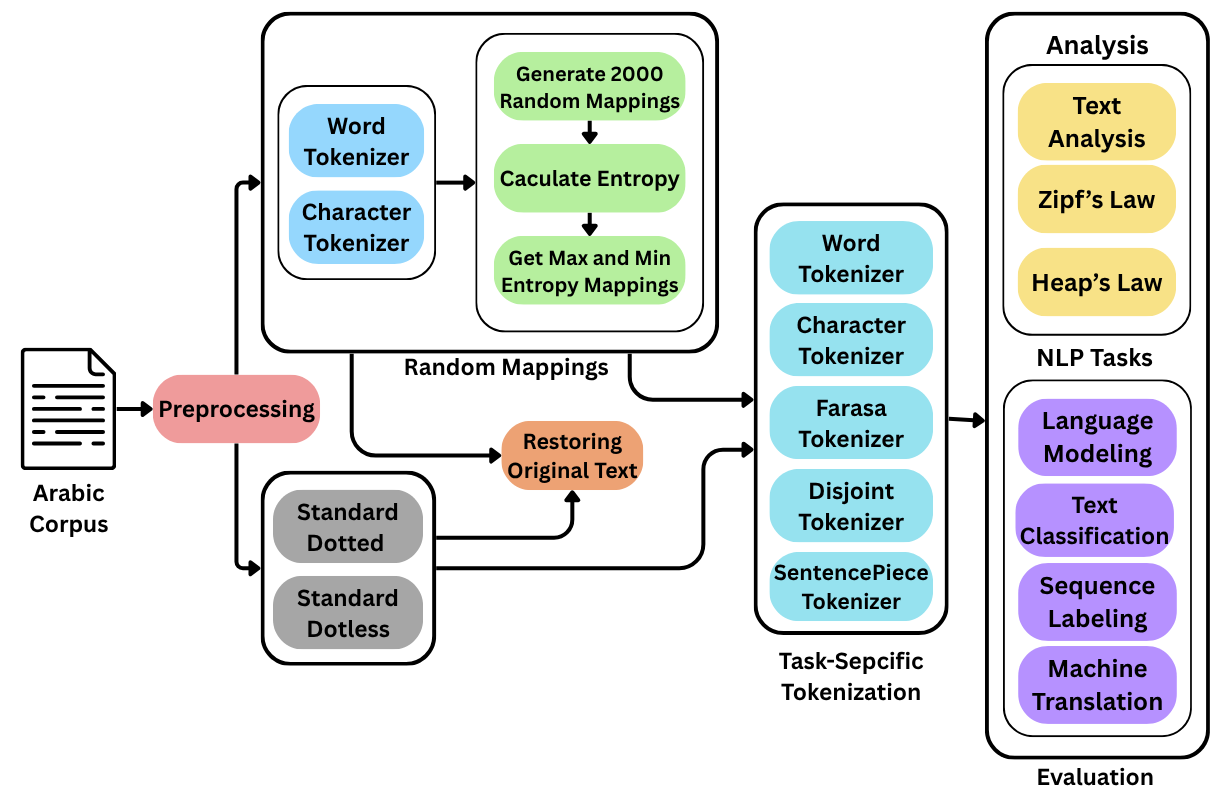}
    \caption{Overview of the methodology used to evaluate random character mappings across NLP tasks.}
    \label{fig:rm-methodology}
\end{figure}

Performance is compared against both the original dotted text and the standard dotless baseline to assess the impact of remapping on linguistic structure and computational efficiency. Figure~\ref{fig:rm-methodology} summarizes the overall workflow for evaluating the performance of random character mappings on various NLP tasks compared to standard dotless mapping, including corpus preprocessing, entropy-based mapping selection, downstream evaluation, and text restoration. 

This section details the experimental framework, random character mapping generation, language analysis, restoration of the original text, language modeling, and downstream task evaluation.

\subsection{Text Tokenization}
\label{tokenization}
Tokenization plays a central role in NLP, as it determines how text is segmented into units processed by models. The choice of tokenizer affects vocabulary size, sequence length, generalization to unseen words, and the ability to capture morphological structure, particularly important for Arabic \citep{alomari2024exploring,alyafeai2023evaluating}.
To ensure a comprehensive evaluation of character remapping, we employ the following tokenization strategies:
\begin{itemize}
    \item  \textbf{Word Tokenizer:} Segments text based on whitespace and punctuation. While it preserves word boundaries, it is sensitive to OOV words and morphological variation.
    
   \item \textbf{Character Tokenizer:} Splits text into individual characters, improving robustness to unseen words but producing longer sequences.
    
    \item \textbf{Farasa Morphological Tokenizer:} Segments words into stems, prefixes, and suffixes, enabling morphology-aware modeling tailored to Arabic.
    
    \item \textbf{Disjoint-Letters Tokenizer:} Leverages the cursive structure of Arabic script by segmenting words at natural letter connection boundaries.
    
    \item \textbf{BPE:} A subword-based tokenizer that learns frequent character sequences, reducing vocabulary size while preserving meaningful units, and commonly used in transformer-based models.
    
\end{itemize}

\begin{figure}[ht]
\centering 
 \includegraphics[width=0.55\linewidth]{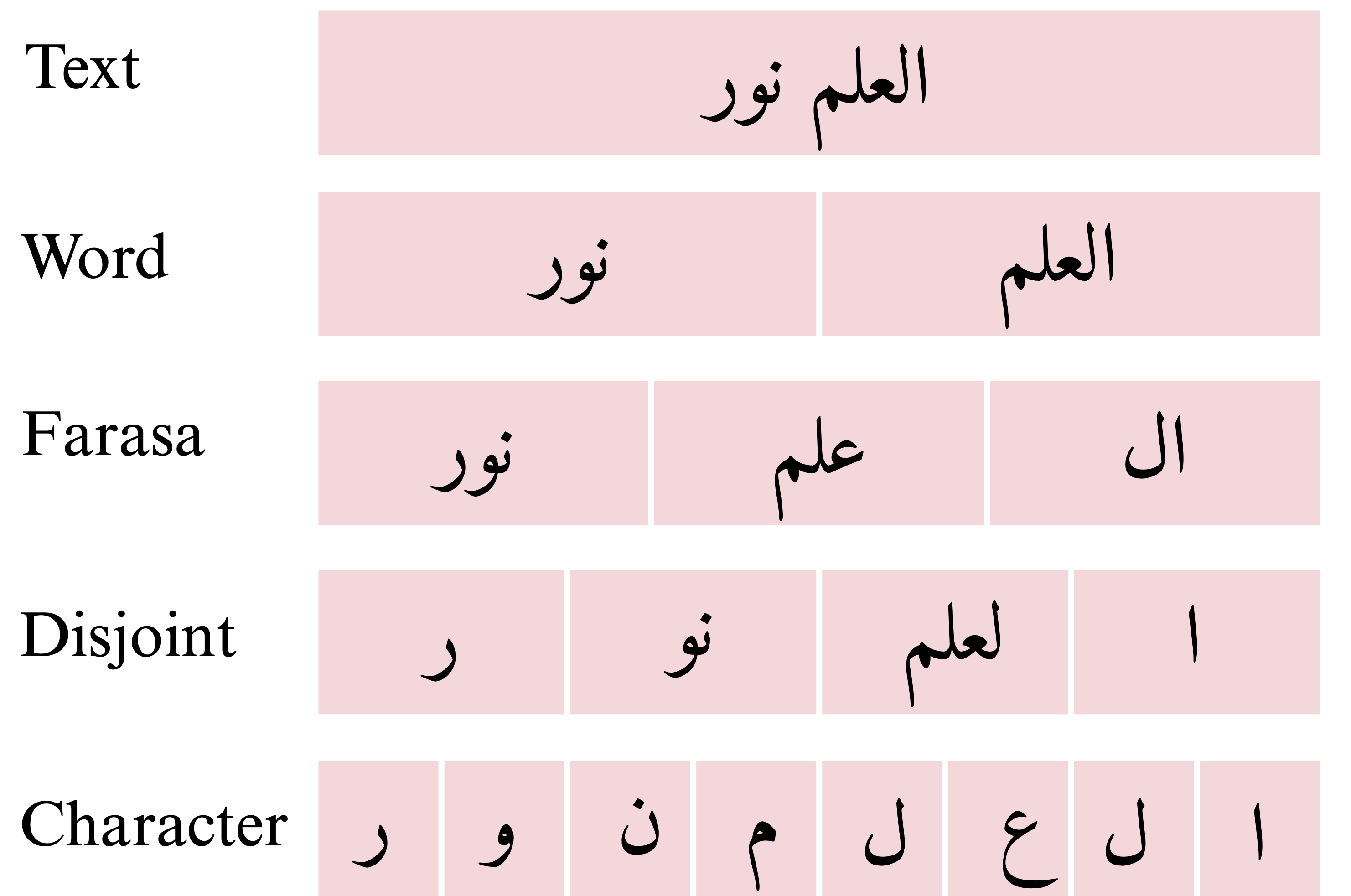} 
\caption{Example of a sentence tokenized using word, farasa, disjoint, and character tokenizers.} 
\label{fig:tokens} 
\end{figure}

An example of these tokenization approaches is provided in Figure~\ref{fig:tokens}.

\subsection{Random Character Mappings}

This study investigates whether character shape and traditional form-pronunciation relationships are essential for NLP performance. Using entropy as a quantitative criterion, we generate and evaluate random character mappings to determine whether models can adapt to arbitrary representations or whether they depend on the original orthographic structure.

Remapping characters directly affects both character-level and word-level entropy by altering the underlying frequency and probability distributions in the text. At the character level, entropy changes according to how predictable the new character distribution becomes. Because words are composed of characters, changes at the regrouping characters naturally propagate to the word level. These alterations affect the vocabulary, as distinct Arabic words may collapse into the same rasm vocabulary, thereby reducing lexical diversity. This grouping reshapes word frequency distributions and, consequently, influences word-level entropy. Higher entropy reflects greater randomness in the linguistic system, while lower entropy indicates increased structural regularity.

By examining the resulting entropy variations, we assess whether NLP models rely on the inherent structure of character forms or whether they can adapt to arbitrary character mappings without substantial performance degradation.


\subsection{Language Analysis}
\label{nalysis}

Statistical regularities are fundamental to language modeling and NLP. Zipf’s Law describes the relationship between word frequency and rank \citep{zipf2016human,masrai2016different}. Zipf’s Law is defined as:
\begin{equation}
    F(w) \propto R(w)^{-\alpha}
\end{equation}
where $R(w)$ denotes the rank of word $w$, $F(w)$ its frequency, and $\alpha$ is a constant typically close to 1.

While Heap’s Law characterizes vocabulary growth with corpus size \citep{heaps1978information,font2013scaling}. Heap’s Law is expressed as:
\begin{equation}
    V = k \times N^{\beta}
\end{equation}
where $V$ represents the vocabulary size (number of unique words), $N$ is the total number of tokens in the corpus, $k$ is a corpus-dependent constant, and $\beta$ is an exponent typically ranging between 0.5 and 1 that governs the rate of vocabulary growth. 

These laws reflect lexical distribution and scaling behavior, which may be affected by changes to the underlying character set.
Shannon entropy provides a core measure of information content by quantifying uncertainty in text \citep{shannon1948mathematical}, defined as:

\begin{equation}
    H = - \sum_{i=1}^{n} P(w_i) \log_2 P(w_i)
\end{equation}

where $P(w_i)$ denotes the probability of token $w_i$ occurring in the corpus.

In script simplification, entropy serves as a principled measure of information loss, as removing dots or remapping characters can alter symbol distributions and increase ambiguity \citep{zitouni2006maximum}. 

We analyze alternative text representations against standard dotted and dotless Arabic by examining vocabulary size, token counts, average token length, and entropy. This allows us to assess the impact of character remapping on linguistic structure, the preservation of statistical regularities (e.g., Zipf’s and Heap’s laws), and overall suitability for NLP applications.

\subsection{Text Restoration}

To ensure human readability and compatibility with downstream applications, we train a Recurrent Neural Network (RNN)-based restoration model to reconstruct the original dotted script from remapped text. The task is framed as a character-level sequence labeling problem, in which each character in the random mapping is assigned its corresponding original dotted form. This enables reversibility of the compression process and supports tasks that require standardized output, such as machine translation. 

\subsection{Language Modeling}

We evaluate alternative representations using a decoder-based transformer language model. The models are trained on randomly mapped text and compared to dotted and dotless baselines. Evaluation considers perplexity, vocabulary size, model size, and training efficiency, enabling us to assess whether random mappings preserve the statistical structure required for effective language modeling.

\subsection{Downstream Tasks}

To evaluate generalization across practical applications, we test the random mappings on multiple NLP tasks. These tasks were selected because they represent widely used NLP benchmarks while probing complementary linguistic capabilities, including semantic understanding, syntactic analysis, and language generation:
\begin{itemize}
    \item \textbf{Text Classification:} This task evaluates the model’s ability to assign a category to a text, including poem meter classification, sentiment analysis, and news topic classification. These benchmarks primarily assess semantic understanding and represent some of the most common NLP applications.
    \item \textbf{Sequence Labeling:} This task assesses token-level predictions, such as part-of-speech (POS) tagging and named entity recognition (NER), measuring the model’s ability to capture grammatical structure and contextual semantic information, respectively.
    \item \textbf{Machine Translation:} This task evaluates the model’s ability to generate accurate translations between Arabic and English under word and subword tokenizations; remapped text is restored to its original form after decoding to ensure translation quality and usability. Translation provides a challenging generation task that requires preserving both semantic meaning and syntactic structure across languages.
\end{itemize}

Together, these experiments provide a comprehensive evaluation of how random character mappings affect both the statistical properties of Arabic and performance across diverse NLP tasks.

\subsection{Evaluation Metrics}
\label{metrics}
We evaluate model performance using perplexity for language modeling, accuracy-based metrics for classification and sequence labeling, and BLEU for machine translation.

Perplexity measures how well a language model predicts a sequence of words \cite{meister2021language}. Lower perplexity indicates stronger predictive performance. It is defined as:

\begin{equation}
\text{PPL}(D, M) = \exp\left(-\frac{1}{N} \sum_{i=1}^{N} \log M(w_i)\right)
\end{equation}

where $N$ is the number of words in dataset $D$, $w_i$ is the $i$-th word, and $M(w_i)$ is the probability assigned by model $M$ to that word.

For classification tasks, we report Accuracy, Precision, Recall, and F1-Score. Accuracy measures the proportion of correctly predicted samples:

\begin{equation}
\text{Accuracy} = \frac{TP + TN}{TP + TN + FP + FN}
\end{equation}

Precision measures the proportion of correctly predicted positive instances:

\begin{equation}
\text{Precision} = \frac{TP}{TP + FP}
\end{equation}

Recall measures the proportion of actual positive instances correctly identified:

\begin{equation}
\text{Recall} = \frac{TP}{TP + FN}
\end{equation}

The F1-Score is the harmonic mean of precision and recall:

\begin{equation}
\text{F1-Score} = \frac{2 \cdot \text{Precision} \cdot \text{Recall}}{\text{Precision} + \text{Recall}}
\end{equation}

where $TP$, $TN$, $FP$, and $FN$ denote true positives, true negatives, false positives, and false negatives.

For machine translation, we use the BLEU score, which evaluates translation quality using modified n-gram precision with a brevity penalty:

\begin{equation}
\text{BLEU} = BP \cdot \exp\left( \sum_{n=1}^{N} w_n \log p_n \right)
\end{equation}

where $p_n$ is the modified precision for n-grams of order $n$, $w_n$ is typically $\frac{1}{N}$, and $N$ is commonly set to 4. The brevity penalty is defined as:

\begin{equation}
BP =
\begin{cases}
1 & \text{if } c > r \\
\exp\left(1 - \frac{r}{c}\right) & \text{if } c \leq r
\end{cases}
\end{equation}

where $c$ is the candidate translation length and $r$ is the reference translation length. Higher BLEU scores indicate closer agreement with reference translations.

\section{Experiments and Results}
\label{sec:Experiments}

\subsection{Experimental Setup}
\label{appendix-a1}
All experiments were conducted on a Linux workstation with two NVIDIA RTX A4500 GPUs (24GB each) using CUDA 12.4. Models were implemented in Python with PyTorch, leveraging GPU acceleration for efficient training and inference. Both RNN-based and transformer-based architectures were evaluated, with task-specific configurations detailed in the corresponding sections.

\subsection{Random Mappings}

\label{preprocessing}
\begin{figure}[ht]
    \centering
    \includegraphics[width=0.85\linewidth]{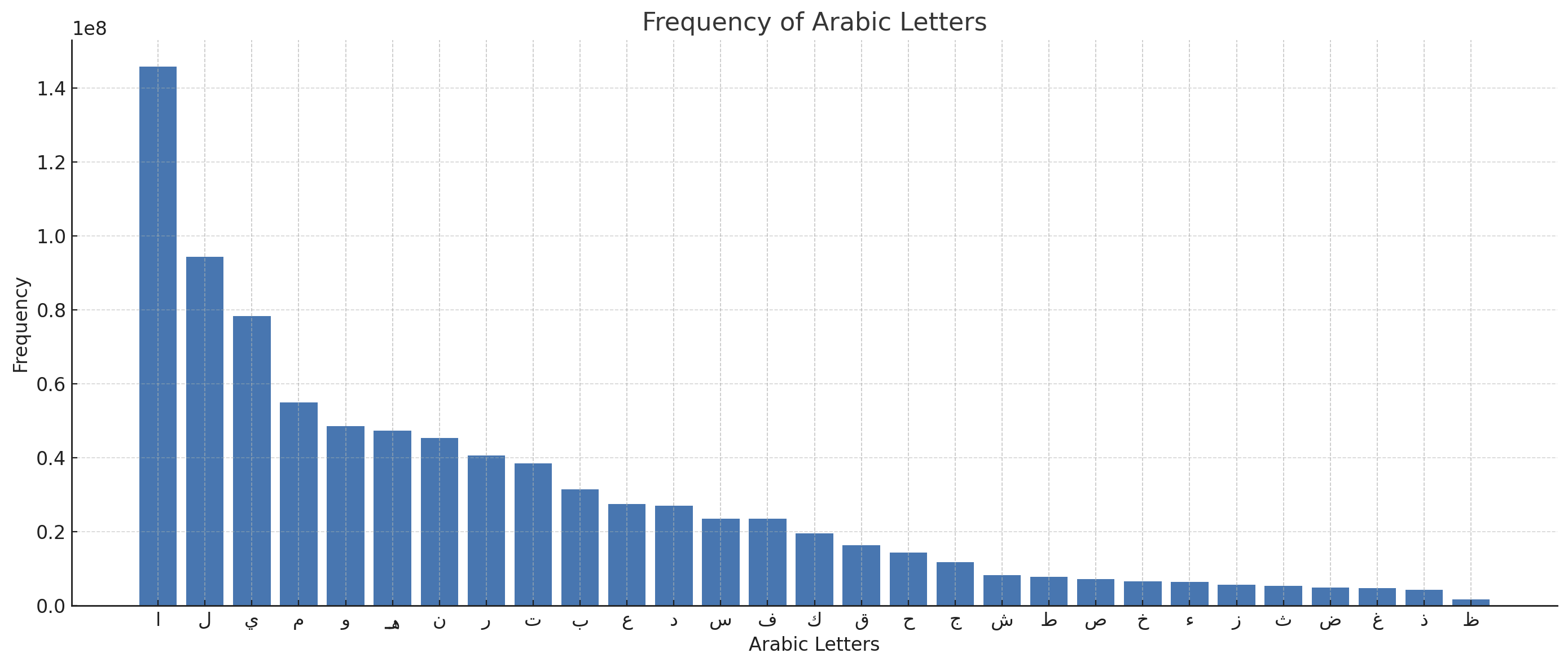}
    \caption{Characters frequency in Wikipedia Arabic corpus.}
    \label{fig:char-freq}
\end{figure}

For generating the random mappings, we used the Wikipedia Arabic Corpus, compiled from Arabic Wikipedia dumps\footnote{\url{https://dumps.wikimedia.org/}} up to October 20, 2022. The full corpus contains over 177 million tokens; due to computational constraints, we relied on a subset of approximately 41 million tokens for the random mapping generation. The corpus covers diverse domains, making it suitable for NLP experiments.
Before applying the character remapping procedures, the text was preprocessed. First, we removed all non-Arabic characters, numbers, and special symbols. We then removed diacritics (tashkil) and elongation marks (tatweel). Finally, we normalized orthographic variants: all forms of alef were unified as (\<ا>), hamza variants were converted to (\<ء>), ta marbuta (\<ة>) was replaced with (\<ه>), and alef maqsura (\<ى>) was replaced with ya (\<ي>). After preprocessing, the character set was reduced to 29 unique Arabic characters, the frequency of the characters is shown in Figure~\ref{fig:char-freq}. 


We conducted the random mapping experiments at both the word and character levels. The word level captures how remapping affects word frequency and structure, while the character level enables a finer-grained analysis of changes in character predictability.

For each strategy, we generated 2{,}000 random character mappings and computed entropy at the corresponding level. From these, we selected four representative mappings: those yielding the Highest Character Entropy (HCE), Lowest Character Entropy (LCE), Highest Word Entropy (HWE), and Lowest Word Entropy (LWE). The selected random mappings are listed in Table~\ref{tab:mappings}, corresponding to the original character rasms and to the mappings that result in the HCE, LCE, HWE, and LWE, respectively.

\begin{table}
\centering
\begin{adjustbox}{max width=\linewidth}
\renewcommand{\arraystretch}{1.5}
\begin{tabular}{|c|c|c|c|c|c|}
\cline{1-6}
\textbf{Rasm} & \textbf{Original} & \textbf{HCE} & \textbf{LCE} & \textbf{HWE} & \textbf{LWE} \\
\cline{1-6}

\<ء> & \<ء> & \<ي> & \<ش> & \<م> & \<ض> \\
\cline{1-6}

\<ا> & \<ا> & \<م> & \<ب> & \<ح> & \<ص> \\
\cline{1-6}
               & \<ب> & \<ج> & \<ا> & \<ث> & \<ر> \\
 \cline{2-6}

\<ٮ> & \<ت> & \<ح> & \<ت> & \<ك> & \<د> \\
\cline{2-6}
               & \<ث> & \<ظ> & \<ل> & \<ض> & \<ف> \\
\cline{1-6}
               & \<ج> & \<خ> & \<و> & \<ز> & \<ي> \\
\cline{2-6}

\<ح> & \<ح> & \<غ> & \<ح> & \<د> & \<ن> \\
\cline{2-6}
               & \<خ> & \<هـ> & \<هـ> & \<خ> & \<ل> \\
\cline{1-6}

\<د> & \<د> & \<ق> & \<م> & \<ط> & \<س> \\
\cline{2-6}
               & \<ذ> & \<ع> & \<ك> & \<ق> & \<ش> \\
\cline{1-6}

\<ر> & \<ر> & \<ص> & \<ن> & \<و> & \<ت> \\
\cline{2-6}
               & \<ز> & \<ض> & \<ر> & \<ء> & \<خ> \\
\cline{1-6}

\<س> & \<س> & \<ذ> & \<ف> & \<ف> & \<و> \\
\cline{2-6}
               & \<ش> & \<ل> & \<ج> & \<ر> & \<ط> \\
\cline{1-6}

\<ص> & \<ص> & \<د> & \<ع> & \<هـ> & \<غ> \\
\cline{2-6}
               & \<ض> & \<ث> & \<غ> & \<ع> & \<هـ> \\
\cline{1-6}

\<ط> & \<ط> & \<ز> & \<ز> & \<ج> & \<ز> \\
\cline{2-6}
               & \<ظ> & \<ر> & \<ي> & \<ي> & \<ب> \\
\cline{1-6}

\<ع> & \<ع> & \<ط> & \<ط> & \<ص> & \<ك> \\
\cline{2-6}
               & \<غ> & \<ء> & \<د> & \<ب> & \<ث> \\
\cline{1-6}

\<ڡ> & \<ف> & \<ن> & \<ض> & \<غ> & \<ج> \\
\cline{1-6}

\<ٯ> & \<ق> & \<ش> & \<ء> & \<ذ> & \<ذ> \\
\cline{1-6}

\<ك> & \<ك> & \<ا> & \<خ> & \<ت> & \<ق> \\
\cline{1-6}

\<ل> & \<ل> & \<ت> & \<ذ> & \<ل> & \<ء> \\
\cline{1-6}

\<م> & \<م> & \<ف> & \<ق> & \<ن> & \<ع> \\
\cline{1-6}

\<ں> & \<ن> & \<و> & \<س> & \<ظ> & \<م> \\
\cline{1-6}

\<هـ> & \<هـ> & \<ك> & \<ث> & \<س> & \<ح> \\
\cline{1-6}

\<و> & \<و> & \<ب> & \<ص> & \<ا> & \<ظ> \\
\cline{1-6}

\<ى> & \<ي> & \<س> & \<ظ> & \<ش> & \<ا> \\
\cline{1-6}

\end{tabular}
\end{adjustbox}
\caption{The character mappings used in this study.}
\label{tab:mappings}
\end{table}

An examination of the selected mappings in Table~\ref{tab:mappings} reveals several noteworthy structural patterns. In some cases, the random mappings retain partial similarities with the standard dotless system. For example, both "\<ر>" and "\<ز>" are mapped to "\<ط>", and "\<ص>" and "\<ض>" share the same mapped form. Likewise, "\<ع>" and "\<غ>" are mapped to "\<ص>", and "\<س>" and "\<ش>" share a common mapping. The character "\<ذ>" is mapped to "\<ٯ>" in both the HWE and LWE configurations.
Although some characters remain mapped to their original rasm, they typically share that rasm with other reassigned characters. A notable exception occurs in the HWE mapping, where "\<ل>" is uniquely mapped to itself without overlap.

More systematic patterns emerge when considering character frequency. In the lowest-entropy mappings, the most frequent characters are collapsed into shared rasms. For instance, "\<ا>", "\<ت>", and "\<ل>" are mapped to "\<ٮ>", while "\<و>", "\<ح>", and "\<هـ>" are mapped to "\<ح>". Similarly, "\<ي>", "\<ن>", and "\<ل>" share the mapping "\<ح>", and "\<ر>", "\<د>", and "\<ف>" are mapped to "\<ٮ>". This consolidation of high-frequency characters reduces distributional diversity and results in lower entropy.

Conversely, in the highest-entropy mappings, frequent characters such as "\<ا>", "\<ي>", "\<م>", and "\<ن>" are assigned distinct rasms, increasing distributional spread. The inverse tendency is observed for low-frequency characters: they are more likely to retain unique rasms in low-entropy mappings, whereas in high-entropy mappings they are often merged.





\subsection{Token Statistics and Entropy Analysis}

We analyze the impact of remapping on token-level statistics and entropy using multiple tokenization strategies. Word-level and character-level tokenizers are discussed below.


\textbf{Word Tokenizer.} Table \ref{tab:rm-analysis-word} summarizes statistics of Wikipedia Arabic corpus representations using a word-level tokenizer, including total tokens (N), vocabulary size (V), vocabulary-to-token ratio (V/N), unique tokens average length, and entropy.

\begin{table}[htbp]
\centering

\begin{adjustbox}{max width=\linewidth}
\begin{tabular}{lccccc}
\hline
\multicolumn{1}{c}{\textbf{Representation}} & 
\multicolumn{1}{c}{\textbf{N}} & 
\multicolumn{1}{c}{\textbf{V}} & 
\multicolumn{1}{c}{\textbf{V/N}} & 
\multicolumn{1}{c}{\textbf{Avg. N}} & 
\multicolumn{1}{c}{\textbf{Entropy}} \\
\hline
Standard Dotted   &\multirow{6}{*}{177,422,512}  & 1,751,650 & 0.0099 & 7.3750  & 13.1701 \\ \cline{1-1} \cline{3-6}
Standard Dotless  && 1,335,665 & 0.0075 & 7.8244 & 12.9327 \\ \cline{1-1} \cline{3-6}
HCE Mapping      && 1,497,015 & 0.0084 & 7.6657 & 13.0173 \\ \cline{1-1} \cline{3-6}
LCE Mapping      && 1,113,525 & 0.0063 & 8.0880  & 12.6694 \\ \cline{1-1} \cline{3-6}
HWE Mapping      && 1,547,034 & 0.0087 & 7.6150  & 13.0663 \\ \cline{1-1} \cline{3-6}
LWE Mapping    && 1,055,960 & 0.0060  & 8.1759 & 12.5047 \\
\hline
\end{tabular}

\end{adjustbox}
\caption{Word tokenizer statistics for Wikipedia Arabic corpus using different text representations.}
\label{tab:rm-analysis-word}
\end{table}

Results show that the standard dotted representation yields the largest vocabulary and highest entropy, reflecting maximum lexical diversity. The standard dotless form reduces vocabulary and entropy due to the merging of letter distinctions.

Among the proposed mappings, LWE and LCE yield the strongest compression, producing lower entropy and smaller vocabularies, as frequent characters are grouped into shared rasm forms. This leads to longer average token lengths, since shorter tokens are merged into the same rasm, reducing the number of unique tokens without a proportional decrease in token length. In contrast, HWE and HCE better preserve lexical diversity, maintaining entropy values closer to the standard dotted form and retaining richer distributions across token forms. This results in shorter average token lengths, as longer tokens are more often merged into shared rasm forms. Overall, entropy positively correlates with vocabulary size: higher entropy reflects greater lexical and length variation, while more aggressive mappings compress tokens and reduce diversity.

\textbf{Character Tokenizer.} Table~\ref{tab:rm-analysis-char} presents character-level statistics for different representations of the Wikipedia Arabic corpus. Because each character is treated as a token, the total token count remains constant across representations, making vocabulary size and entropy the main distinguishing metrics.

\begin{table}[htbp]
\centering

\begin{adjustbox}{max width=\linewidth}
\begin{tabular}{p{3.5cm}ccc}
\hline
\textbf{Representation} & \textbf{N} & \textbf{V} & \textbf{Entropy} \\
\hline
Standard Dotted   &\multirow{6}{*}{851,699,872} & 29 & 4.1903 \\ \cline{1-1} \cline{3-3} \cline{4-4}
Standard Dotless  & &\multirow{5}{*}{19} & 3.8920 \\ \cline{1-1} \cline{4-4}
HCE Mapping     &&& 3.9233 \\ \cline{1-1} \cline{4-4}
LCE Mapping     &&& 3.2435 \\ \cline{1-1} \cline{4-4}
HWE Mapping       &&& 3.7849 \\ \cline{1-1} \cline{4-4}
LWE Mapping     &&& 3.4580 \\ \cline{1-1} \cline{4-4}
\hline
\end{tabular}
\end{adjustbox}
\caption{Character tokenizer statistics for Wikipedia Arabic corpus using different text representations.}
\label{tab:rm-analysis-char}
\end{table}

The standard dotted representation retains 29 unique characters and achieves the highest entropy, reflecting the full range of orthographic distinctions. All other representations reduce the character set to 19 symbols by merging dotted variants, confirming the intended structural simplification.

Entropy clearly captures the degree of compression. The standard dotless form lowers entropy due to character merging. HCE maintains a relatively higher entropy, followed by HWE, indicating moderate preservation of variability. In contrast, LCE and LWE produce the lowest entropy values, reflecting stronger character-level compression and increased repetition.

Analysis of the Farasa and Disjoint tokenization techniques is provided in Tables~\ref{tab:rm-analysis-farasa} and \ref{tab:rm-analysis-disjoint} in Appendix~B, revealing patterns consistent with those observed in previous techniques.

\subsection{Zipf’s and Heap’s Law Analysis}

 To evaluate whether alternative character mappings preserve the statistical properties captured by Zipf’s and Heap’s laws, we analyze six representations, standard dotted, standard dotless, and four alternative mappings, using the Arabic Wikipedia corpus under word-level tokenization.

\begin{figure}[ht]
    \centering
    \includegraphics[width=0.8\linewidth]{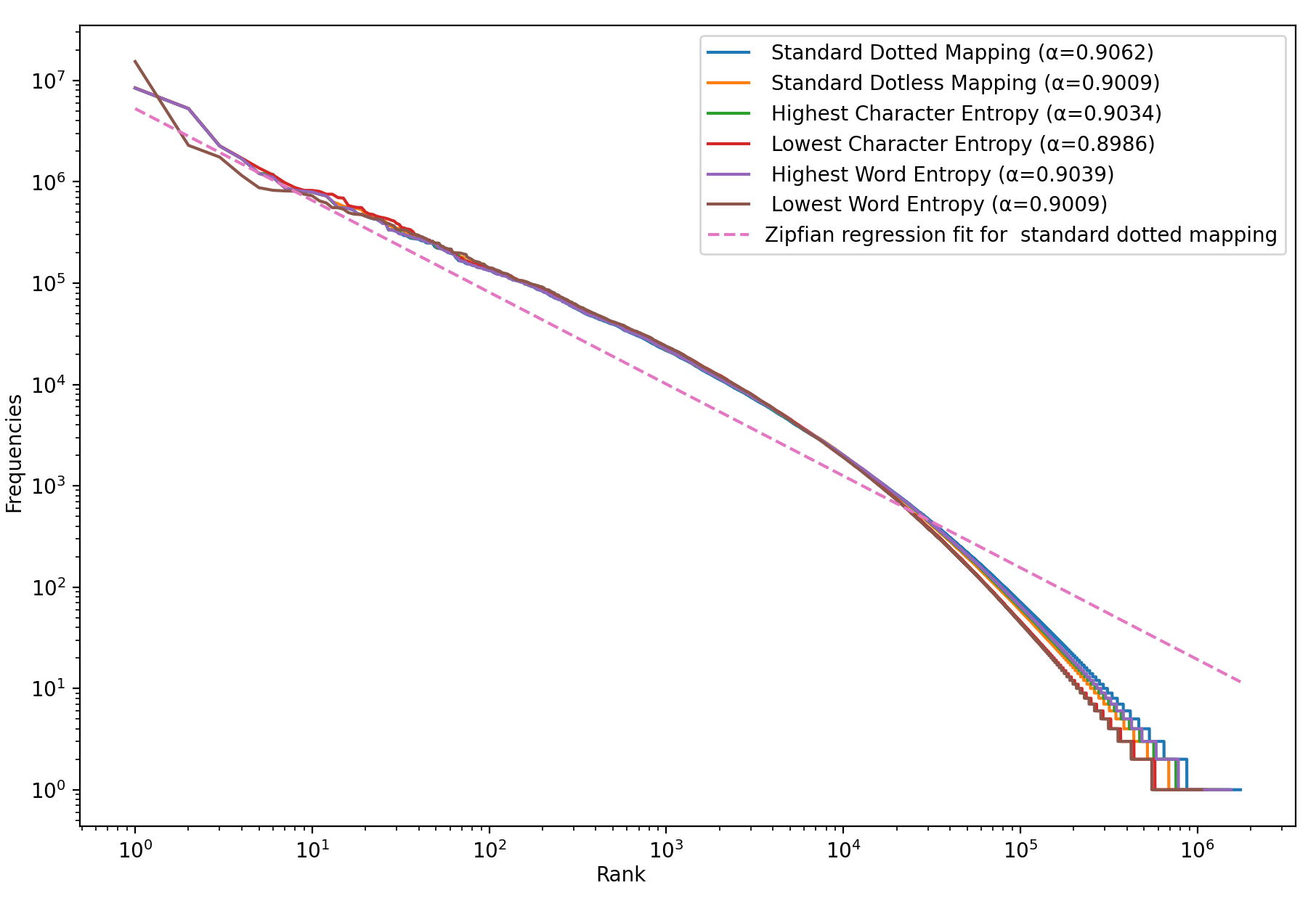}
    \caption{Zipf’s Law plot under word tokenizer for various character mappings.}
    \label{fig:rm-zipfs-word}
\end{figure}

Figure~\ref{fig:rm-zipfs-word} shows that all mappings closely follow Zipfian behavior, with $\alpha$ values ranging from 0.8986 to 0.9062. The rank-frequency curves largely overlap, indicating that character remapping has minimal impact on global frequency distributions at the word level. The standard dotted and dotless forms are nearly identical, with minor differences in the tail. The HCE and HWE mappings extend slightly further in the long tail, while the LWE and LCE mappings exhibit marginally steeper decay, though overall Zipfian structure is preserved.

\begin{figure}[ht]
    \centering
    \includegraphics[width=0.8\linewidth]{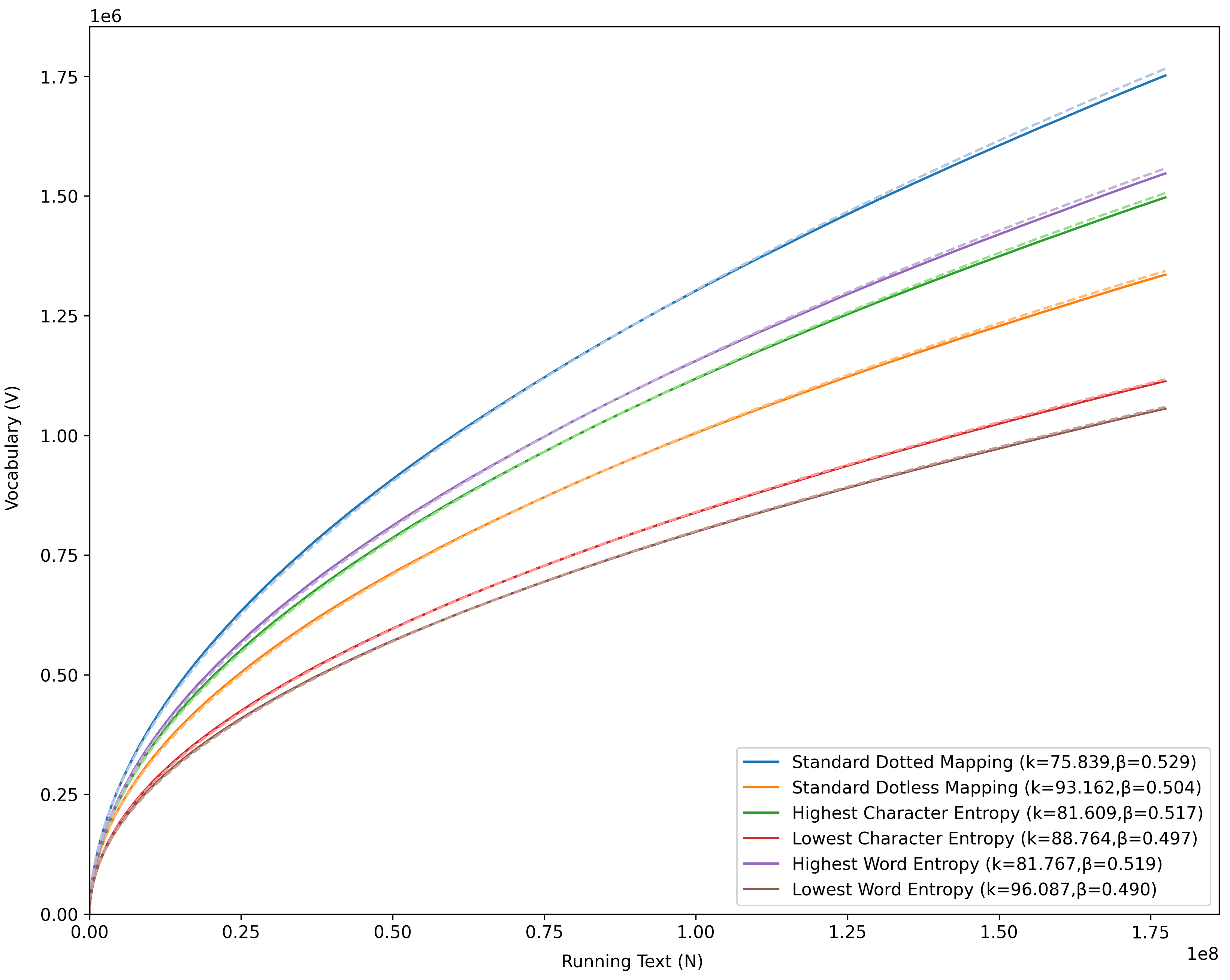}
    \caption{Heap’s Law plot under word tokenizer for various character mappings,
(dashed lines represent the regression fit for each representation).}
    \label{fig:rm-heaps-word}
\end{figure}

Figure~\ref{fig:rm-heaps-word} illustrates vocabulary growth, where all mappings follow Heap’s Law with $\beta$ values between 0.490 and 0.529, confirming stable power-law behavior. The standard dotted representation yields the highest vocabulary growth, reflecting greater lexical diversity, while the dotless form grows more slowly due to character merging. The HWE and HCE mappings fall between these extremes, preserving more diversity than the dotless baseline. In contrast, the LWE and LCE mappings show the slowest growth, with LWE exhibiting the strongest compression. 

Zipf's and Heap's analysis of the Farasa and Disjoint tokenization techniques is provided in Figures~\ref{fig:rm-zipfs-farasa}--\ref{fig:rm-heaps-disjoint} in Appendix~B, highlighting trends that align with those observed in previous techniques.

\subsection{Restoring Original Mapping}
\label{rm-re-experiments}

We evaluate whether the standard dotted Arabic script can be accurately restored from remapped text using a neural sequence model. 
The restoration model was evaluated only on the standard dotless and LCE mappings. The dotless mapping serves as the baseline, while LCE represents a strongly compressed low-entropy mapping with competitive performance on the downstream machine translation task.
The task is formulated as a character-level sequence labeling problem, where each character in the alternative representation is mapped back to its original dotted form. Experiments are conducted on the Arabic Wikipedia corpus using character-level tokenization with standard preprocessing. The restoration is performed using a stacked Bidirectional LSTM (BiLSTM) model; details are provided in Table~\ref{tab:model-architecture} in Appendix A.



 Performance is evaluated using Word Error Rate (WER) and Character Error Rate (CER). Two input mappings are tested: the standard dotless form and the LCE mapping. Results show strong recovery performance in both cases:

\begin{itemize}
    \item \textbf{Dotless:} WER = 0.0510, CER = 0.0125
    \item \textbf{LCE:} WER = 0.0796, CER = 0.0235
\end{itemize}

Although restoration from the dotless form yields lower error rates, performance on the more compressed LCE mapping remains robust. Some examples of the restored text are shown in Table~\ref{tab:restored-samples}. Overall, the low error rates indicate that the original orthography can be reliably reconstructed even from heavily remapped text, supporting the practical usability of such transformations in downstream NLP tasks.

\begin{table}[htbp]
\centering
\begin{adjustbox}{max width=\linewidth}
\begin{tabular}{|p{5.2cm}|p{5.2cm}|p{4.5cm}|}
\hline
\textbf{Original Text (Dotted)} & \textbf{Input (Mapped)} & \textbf{Restored Output} \\
\hline
\shortstack[r]{
\RL{ وهنالك من يعتقد بأن الراحة ما}\\
\RL{ هي إلا وقت فائض عن الحاجة}\\
\RL{ بعد انقضاء ضرورات الحياة}
}
&
\shortstack[r]{
\RL{ححرٮٮد در طصٮمع اٮر ٮٮرٮحح دٮ}\\
\RL{  حط ٮٮٮ حمٮ سٮڡڡ صر ٮٮحٮسح}\\
\RL{ اصع ٮرمڡٮٯ ڡرحرٮٮ ٮٮحطٮح}
}
&
\shortstack[r]{
\RL{وهنالك من يعتقد بأن الراوح ما}\\
\RL{ هي إلا حقا فائض عن الحاجة}\\
\RL{  بعد انقضاء ضرورات الحياة}
}
\\
\hline
\RL{انتخب عضو مجلس النواب من بلجيكا}
&
\RL{ٮرٮكا صڡح دسٮں ٮٮرحٮا در اٮسطدٮ}
&
\RL{انتخب عضو مجلس النواب من بلجيكا}
\\
\hline
\end{tabular}
\end{adjustbox}
\caption{Examples of original and restored text representation (from the LCE mapping) using the character restoration model.}
\label{tab:restored-samples}
\end{table}

\subsection{Language Modeling}
\label{lm-experiments}
We investigate the impact of character remapping on transformer-based language modeling. Experiments on the Arabic Wikipedia corpus evaluate how different text representations affect vocabulary size, OOV rates, and perplexity.

The corpus is cleaned, normalized, and processed using multiple tokenization strategies. To control vocabulary imbalance, the vocabulary is limited to cover 95\% of token frequency, with the remaining tokens mapped to <UNK>. A causal Transformer language model is trained with an autoregressive next-token prediction objective. Hyperparameters are kept consistent across all representations to ensure a fair comparison. The hyperparameter configurations are provided in Table~\ref{tab:transformer-model} in Appendix~A. Performance is evaluated using perplexity.

The resulting vocabulary sizes and OOV rates of the language model on the test set under word-level tokenization are reported in Table~\ref{tab:oov-table-word}.

\begin{table}[htbp]
\centering
\renewcommand{\arraystretch}{1.3}

\begin{adjustbox}{max width=\linewidth}
\begin{tabular}{lccc}
\hline
\textbf{Text Type} & \textbf{N} & \textbf{OOVs} & \textbf{OOV\%} \\
\hline
Standard Dotted   & \multirow{6}{*}{6,244,635} & 318,936 & 5.11 \\ \cline{1-1} \cline{3-4}
Standard Dotless  &  & 316,877 & 5.07 \\  \cline{1-1} \cline{3-4}
HCE Mapping      &  & 317,285 & 5.08 \\ \cline{1-1} \cline{3-4}
LCE Mapping       &  & 314,646 & 5.04 \\ \cline{1-1} \cline{3-4}
HWE Mapping      &  & 317,765 & 5.09 \\ \cline{1-1} \cline{3-4}
LWE Mapping       &  & 314,984 & 5.04 \\
\hline
\end{tabular}
\end{adjustbox}
\caption{OOV statistics for different text representations using the word tokenizer.}
\label{tab:oov-table-word}
\end{table}

The standard dotless representation slightly reduces the OOV rate compared to the dotted baseline, indicating a modest benefit from removing dots. More substantially, the LCE and LWE mappings achieve the lowest OOV rate (5.04\%), suggesting that reducing character-set complexity improves generalization by decreasing unseen word forms. In contrast, the HCE and HWE mappings provide little improvement over the baseline.

Table \ref{tab:rm-lm-vocab-perplexity} shows that the standard dotted representation generally achieves the lowest perplexity, reflecting its rich lexical distinctions, but at the cost of the largest vocabulary. Among alternative mappings, HWE performs competitively across tokenizers and even matches or surpasses the baseline in certain settings, demonstrating that partial lexical preservation can maintain modeling quality. LCE and LWE substantially reduce vocabulary size (up to nearly 50\%) with only moderate increases in perplexity, indicating an efficient trade-off between compression and performance.

\begin{table}[htbp]
\centering
\renewcommand{\arraystretch}{1.3}

\begin{adjustbox}{max width=\linewidth}
\begin{tabular}{lrcrcrrcc}
\hline
\multicolumn{1}{c}{\textbf{Text Type}} & 
\multicolumn{2}{c|}{\textbf{Word}} & 
\multicolumn{2}{c|}{\textbf{Farasa}} & 
\multicolumn{2}{c|}{\textbf{Disjoint}} & 
\multicolumn{2}{c}{\textbf{Character}} \\
\cline{2-9}
 & \multicolumn{1}{c}{\textbf{V}} & 
   \multicolumn{1}{c}{\textbf{PPL}} & 
   \multicolumn{1}{c}{\textbf{V}} & 
   \multicolumn{1}{c}{\textbf{PPL}} & 
   \multicolumn{1}{c}{\textbf{V}} & 
   \multicolumn{1}{c}{\textbf{PPL}} & 
   \multicolumn{1}{c}{\textbf{V}} & 
   \multicolumn{1}{c}{\textbf{PPL}} \\
\hline
Standard Dotted & 98,970 & 191.00 & 44,833 & 8.25 & 23,961 & 6.50 & 29 & 3.19 \\ \hline
Standard Dotless & 70,684 & 201.27 & 27,095 & 8.39 & 12,568 & 6.83 & \multirow{5}{*}{19} & 3.21 \\ \cline{1-7} \cline{9-9}
HCE Mapping & 79,228 & 200.46 & 32,137 & 8.24 & 23,961 & 8.12 &  & 3.25 \\ \cline{1-7} \cline{9-9}
LCE Mapping & 52,684 & 207.09 & 19,558 & 8.49 & 14,090 & 9.15 &  & 3.23 \\ \cline{1-7} \cline{9-9}
HWE Mapping & 83,481 & 200.28 & 34,147 & 8.25 & 15,738 & 6.28 &  & 3.21 \\ \cline{1-7} \cline{9-9}
LWE Mapping & 50,708 & 207.36 & 17,194 & 8.58 & 25,646 & 11.91 &  & 3.19 \\
 \hline
\end{tabular}
\end{adjustbox}
\caption{Vocabulary size and perplexity of language models for different tokenizers and text representations.}
\label{tab:rm-lm-vocab-perplexity}
\end{table}

Tokenizer choice significantly influences outcomes. Farasa segmentation generally lowers perplexity and reduces performance gaps between mappings, while the disjoint tokenizer is more sensitive to character-level differences. In contrast, character-level tokenization shows nearly uniform perplexity across representations, suggesting minimal impact from remapping at this granularity.


Table~\ref{tab:rm-lm-modelsize-time} shows that the standard dotted representation yields the largest models and longest training times due to its large vocabulary. All alternative mappings reduce both model size and training time, with the LWE and LCE mappings achieving the greatest efficiency gains while incurring only modest increases in perplexity. Even the highest entropy mappings improve efficiency compared to the dotted baseline. Overall, character remapping provides a clear trade-off: substantial reductions in model size and training cost with limited impact on language modeling performance.

\begin{table}[htbp]
\centering
\renewcommand{\arraystretch}{1.3}
\begin{adjustbox}{max width=0.95\linewidth}
\begin{tabular}{lrrrrrrrr}
\hline
\multicolumn{1}{c}{\textbf{Text Type}} & 
\multicolumn{2}{c|}{\textbf{Word}} & 
\multicolumn{2}{c|}{\textbf{Farasa}} & 
\multicolumn{2}{c|}{\textbf{Disjoint}} & 
\multicolumn{2}{c}{\textbf{Character}} \\
\cline{2-9}
& \multicolumn{1}{c}{\textbf{S}} & 
  \multicolumn{1}{c}{\textbf{T}} & 
  \multicolumn{1}{c}{\textbf{S}} & 
  \multicolumn{1}{c}{\textbf{T}} & 
  \multicolumn{1}{c}{\textbf{S}} & 
  \multicolumn{1}{c}{\textbf{T}} & 
  \multicolumn{1}{c}{\textbf{S}} & 
  \multicolumn{1}{c}{\textbf{T}} \\
\hline
Standard Dotted     & 110.00 & 7466.72 & 55.40 & 7995.69 & 34.00 & 7299.14 & 9.50 & 5107.63 \\ \hline
Standard Dotless    & 81.90  & 5755.92 & 37.20 & 5909.86 & 22.30 & 5562.38 & 9.50 & 5110.21 \\ \hline
HCE Mapping        & 90.70  & 6279.64 & 42.40 & 6500.16 & 34.10 & 7310.36 & 9.50 & 5114.20 \\ \hline
LCE Mapping         & 63.50  & 4693.05 & 29.00 & 5141.76 & 23.90 & 5803.97 & 9.50 & 5121.17 \\ \hline
HWE Mapping        & 95.00  & 6533.07 & 44.50 & 6793.74 & 25.60 & 6066.78 & 9.50 & 5136.49 \\ \hline
LWE Mapping         & 61.40  & 4509.84 & 27.10 & 4854.49 & 35.70 & 7505.11 & 9.50 & 5110.01 \\
\hline
\end{tabular}
\end{adjustbox}
\caption{Model size (S) in MB and average training time per epoch (T) in seconds
across different text representations and tokenizers for language modeling.}
\label{tab:rm-lm-modelsize-time}
\end{table}

Overall, the findings demonstrate that language models can still capture key statistical structure even under arbitrary mappings while reducing vocabulary size and OOV rates.



\subsection{Poem Meter Classification}  
This task focuses on recognizing metrical patterns in classical Arabic poetry, making it highly sensitive to character-level and phonological features. Experiments are conducted on the Ashaar dataset \citep{alyafeai2023ashaar}, containing over 897,000 verses filtered to include only the 16 canonical meter classes. Character-level tokenization is applied in line with prior studies \citep{abandah2022classifying}. A Gated Recurrent Unit (GRU)-based model was used for this task, configurations are shown in Table~\ref{tab:pm_model_config} in Appendix~A.



As shown in Table \ref{tab:rm-poetry-meter-character-results}, the standard dotted representation achieves the best performance, benefiting from full orthographic detail. However, HCE and HWE mappings achieve closely competitive results while reducing the character set from 29 to 19 symbols. The dotless, LCE and LWE mappings show a modest performance drop but remain competitive. Overall, random mappings demonstrate robustness even in rhythm-sensitive, character-dependent tasks like poem meter classification, achieving high accuracy while offering an effective trade-off between reduced character set complexity and performance.

\begin{table}[htbp]
\centering
\renewcommand{\arraystretch}{1.3}

\begin{adjustbox}{max width=\linewidth}
\begin{tabular}{lcccc}
\hline
\multicolumn{1}{c}{\textbf{Text Type}} & 
\multicolumn{1}{c}{\textbf{Precision}} & 
\multicolumn{1}{c}{\textbf{Recall}} & 
\multicolumn{1}{c}{\textbf{F1-score}} & 
\multicolumn{1}{c}{\textbf{Accuracy}} \\
\hline

Standard Dotted    & 0.925 & 0.913 & 0.919 & 0.960 \\ \hline
Standard Dotless  & 0.865 & 0.867 & 0.864 & 0.946 \\ \hline
HCE Mapping      & 0.910 & 0.906 & 0.908 & 0.955 \\ \hline
LCE Mapping      & 0.832 & 0.852 & 0.838 & 0.930 \\ \hline
HWE Mapping      & 0.905 & 0.914 & 0.908 & 0.955 \\ \hline
LWE Mapping     & 0.799 & 0.838 & 0.811 & 0.927 \\ 
\hline
\end{tabular}
\end{adjustbox}
\caption{The results of poem meter classification for different text representation using character tokenizer.}
\label{tab:rm-poetry-meter-character-results}
\end{table}

\subsection{Sentiment Analysis}
\label{sa-experiments}

This experiment evaluates the impact of character remapping on binary sentiment classification. We use the LABR dataset \citep{aly2013labr}, which contains 63,257 Arabic book reviews. The task is reformulated as binary classification: ratings 1-2 as negative and 4-5 as positive, resulting in 9,225 training samples and 2,338 testing samples. Text preprocessing follows the same normalization pipeline used in previous experiments. A GRU-based neural model is employed, details are shown in Table~\ref{tab:sa_hyperparams} in Appendix A.


The results reported in Table \ref{tab:rm-classification-vocab-accuracy} indicate that sentiment classification remains stable across representations, despite substantial vocabulary reductions. Under word tokenization, HWE and HCE mappings achieve the highest or near-highest accuracy while reducing vocabulary size by 10-14\%. LCE and LWE mappings reduce vocabulary by over 35\% with only minor accuracy drops. With Farasa tokenization, overall accuracy improves. Under disjoint tokenization, performance slightly decreases but similar trends persist. 

\begin{table}[htbp]
\centering
\renewcommand{\arraystretch}{1.3}

\begin{adjustbox}{max width=\linewidth}
\begin{tabular}{lrcrcrc}
\hline
\multicolumn{1}{c}{\textbf{Text Type}} & 
\multicolumn{2}{c|}{\textbf{Word}} & 
\multicolumn{2}{c|}{\textbf{Farasa}} & 
\multicolumn{2}{c}{\textbf{Disjoint}} \\
\cline{2-7}
& \multicolumn{1}{c}{\textbf{V}} & 
  \multicolumn{1}{c}{\textbf{Acc.}} & 
  \multicolumn{1}{c}{\textbf{V}} & 
  \multicolumn{1}{c}{\textbf{Acc.}} & 
  \multicolumn{1}{c}{\textbf{V}} & 
  \multicolumn{1}{c}{\textbf{Acc.}} \\
\hline

Standard Dotted   & 23,404 & 0.759 & 12,745 & 0.776 & 8,998  & 0.754 \\ \hline
Standard Dotless  & 17,548 & 0.760 & 8,337  & 0.770 & 5,147  & 0.756 \\\hline
HCE Mapping     & 20,151 & 0.761 & 10,120 & 0.768 & 7,723  & 0.754 \\\hline
LCE Mapping     & 14,766 & 0.758 & 6,981  & 0.781 & 5,277  & 0.759 \\\hline
HWE Mapping       & 20,982 & 0.768 & 10,649 & 0.781 & 5,979  & 0.766 \\\hline
LWE Mapping     & 14,666 & 0.746 & 6,462  & 0.776 & 8,280  & 0.735 \\
\hline
\end{tabular}
\end{adjustbox}
\caption{Accuracy and vocabulary size of sentiment analysis for different text
representation using various tokenizer.}
\label{tab:rm-classification-vocab-accuracy}
\end{table}

Table~\ref{tab:rm-classification-model-training} shows that alternative character mappings substantially reduce model size and training time in sentiment classification while maintaining competitive accuracy. The LCE and LWE mappings consistently produce the smallest and fastest models, despite using vocabularies over 35\% smaller than the standard dotted representation, and achieve comparable performance under Farasa. The HWE mapping offers a strong balance between efficiency and accuracy, often reaching the best results while still reducing model size relative to the baseline. Overall, the results demonstrate that compressed mappings enable lighter and faster models with minimal loss in classification performance, making them well-suited for resource-constrained settings.

\begin{table}[htbp]
\centering
\renewcommand{\arraystretch}{1.3}

\begin{adjustbox}{max width=\linewidth}
\begin{tabular}{lrrrrrr}
\hline
\multicolumn{1}{c}{\textbf{Text Type}} & 
\multicolumn{2}{c|}{\textbf{Word}} & 
\multicolumn{2}{c|}{\textbf{Farasa}} & 
\multicolumn{2}{c}{\textbf{Disjoint}} \\
\cline{2-7}
& \multicolumn{1}{c}{\textbf{S}} & 
  \multicolumn{1}{c}{\textbf{T}} & 
  \multicolumn{1}{c}{\textbf{S}} & 
  \multicolumn{1}{c}{\textbf{T}} & 
  \multicolumn{1}{c}{\textbf{S}} & 
  \multicolumn{1}{c}{\textbf{T}} \\
\hline

Standard Dotted   & 12.3 & 9.388 & 4.1 & 11.963 & 3.5  & 15.811 \\ \hline
Standard Dotless  & 9.3  & 11.371 & 2.9 & 12.923 & 2.5  & 19.157 \\ \hline
HCE Mapping      & 10.7 & 12.652 & 3.4 & 13.866 & 3.2  & 19.321 \\ \hline
LCE Mapping       & 7.9  & 13.167 & 2.6 & 14.809 & 2.5  & 19.806 \\ \hline
HWE Mapping      & 11.1 & 14.338 & 3.5 & 15.724 & 2.7  & 18.706 \\ \hline
LWE Mapping       & 7.9  & 9.290  & 2.4 & 16.561 & 3.3  & 13.815 \\
\hline
\end{tabular}
\end{adjustbox}
\caption{Model size (S) in MB and average training time per epoch (T) in seconds
across different text representations and tokenizers for sentiment analysis.}
\label{tab:rm-classification-model-training}
\end{table}
In summary, the results indicate that sentiment information is preserved under arbitrary mappings, enabling competitive classification performance across multiple tokenization strategies with a substantially reduced vocabulary.

\subsection{News Topic Classification}
\label{nt-experiments}

This experiment evaluates the effect of character remapping on multi-class news topic classification. We use the Sanad dataset (AlKhaleej subset) \citep{einea2019sanad}, which contains 40,950 training and 4,550 testing articles labeled across seven categories: culture, economy, health, politics, religion, sports, and technology. Standard normalization and cleaning procedures are applied, consistent with previous experiments. Classification is performed using a BiGRU-based model, hyperparameters are shown in Table~\ref{table:nt_model_config} in Appendix A. Accuracy is used as the evaluation metric.


Table \ref{tab:rm-topic-classification-vocab-accuracy} shows consistently high performance across all mappings and tokenizers (accuracy generally between 0.973 and 0.981), indicating strong robustness to character remapping. Under word tokenization, LCE and HWE achieve the highest accuracy while reducing vocabulary size by up to 38\%. Farasa tokenization yields stable and tightly clustered results. Under the disjoint tokenizer, LWE achieves the highest accuracy.

\begin{table}[htbp]
\centering

\renewcommand{\arraystretch}{1.3}

\begin{adjustbox}{max width=\linewidth}
\begin{tabular}{lrcrcrc}
\hline
\multicolumn{1}{c}{\textbf{Text Type}} & 
\multicolumn{2}{c|}{\textbf{Word}} & 
\multicolumn{2}{c|}{\textbf{Farasa}} & 
\multicolumn{2}{c}{\textbf{Disjoint}} \\
\cline{2-7}
& \multicolumn{1}{c}{\textbf{V}} & 
  \multicolumn{1}{c}{\textbf{Acc.}} & 
  \multicolumn{1}{c}{\textbf{V}} & 
  \multicolumn{1}{c}{\textbf{Acc.}} & 
  \multicolumn{1}{c}{\textbf{V}} & 
  \multicolumn{1}{c}{\textbf{Acc.}}  \\
\hline

Standard Dotted   & 32,402 & 0.971 & 15,430 & 0.978 & 10,158 & 0.975 \\ \hline
Standard Dotless  & 24,978 & 0.973 & 10,585 & 0.977 & 5,911  & 0.978 \\ \hline
HCE Mapping     & 27,446 & 0.976 & 12,162 & 0.975 & 10,003 & 0.979 \\ \hline
LCE Mapping     & 20,028 & 0.980 & 8,576  & 0.977 & 6,810  & 0.973  \\ \hline
HWE Mapping       & 28,748 & 0.978 & 12,835 & 0.975 & 6,846  & 0.975 \\ \hline
LWE Mapping     & 19,540 & 0.976 & 7,720  & 0.976 & 10,693 & 0.981  \\
\hline
\end{tabular}
\end{adjustbox}
\caption{Accuracy and vocabulary size of news topic classification for different text representation using various tokenizer.}
\label{tab:rm-topic-classification-vocab-accuracy}
\end{table}


Table~\ref{tab:rm-topic-classification-model-training} shows that alternative character mappings substantially reduce model size and training time in topic classification while maintaining nearly identical accuracy to the standard dotted baseline. In particular, the LCE and LWE mappings reduce model size by up to 38\% and shorten training time, despite large drops in vocabulary size, with accuracy remaining stable. Overall, topic classification is relatively insensitive to character remapping; substantial vocabulary compression yields lighter and faster models with minimal impact on performance, highlighting the robustness and efficiency of alternative mappings across tokenization strategies.

\begin{table}[htbp]
\centering
\renewcommand{\arraystretch}{1.3}
\begin{adjustbox}{max width=\linewidth}
\begin{tabular}{lrrrrrr}
\hline
\multicolumn{1}{c}{\textbf{Text Type}} & 
\multicolumn{2}{c|}{\textbf{Word}} & 
\multicolumn{2}{c|}{\textbf{Farasa}} & 
\multicolumn{2}{c}{\textbf{Disjoint}}\\
\cline{2-7}
& \multicolumn{1}{c}{\textbf{S}} & 
  \multicolumn{1}{c}{\textbf{T}} & 
  \multicolumn{1}{c}{\textbf{S}} & 
  \multicolumn{1}{c}{\textbf{T}} & 
  \multicolumn{1}{c}{\textbf{S}} & 
  \multicolumn{1}{c}{\textbf{T}} \\
\hline

Standard Dotted   & 4.6 & 29.972 & 8.7 & 270.052 & 3.1 & 197.293  \\ \hline
Standard Dotless  & 3.7 & 31.478 & 8.1 & 272.690 & 2.5 & 199.195  \\ \hline
HCE Mapping      & 4.0 & 32.646 & 8.3 & 273.299 & 3.1 & 181.229  \\ \hline
LCE Mapping       & 3.1 & 33.435 & 7.8 & 273.683 & 2.6 & 177.415 \\ \hline
HWE Mapping      & 4.2 & 34.530 & 8.3 & 274.450 & 2.7 & 212.807 \\ \hline
LWE Mapping       & 3.0 & 29.296 & 7.7 & 274.876 & 3.1 & 148.561  \\
\hline
\end{tabular}
\end{adjustbox}
\caption{Model size (S) in MB and average training time per epoch (T) in seconds
for news topic classification using different text representations and tokenizers.}
\label{tab:rm-topic-classification-model-training}
\end{table}

Overall, news topic classification performance remains highly stable under remapping, demonstrating that semantic category recognition is largely preserved even with random character mappings.

\subsection{Part-of-Speech Tagging}
\label{pos-experiments}

This experiment evaluates the impact of character remapping on Arabic POS tagging. Experiments are conducted on the Arabic subset of the PADT from the Universal Dependencies project \citep{smrz2008prague}, consisting of 5,959 training, 906 validation, and 674 testing sentences annotated with 17 POS tags (plus one padding tag). A BiLSTM-based model is used, and hyperparameters are fixed across all mappings for fair comparison, as shown in Table~\ref{tab:pos_model_config} in Appendix A. Performance is evaluated using accuracy as the primary metric, along with precision, recall, and F1-score to account for class imbalance.



Table \ref{tab:rm-pos-word-results} indicates that the standard dotted representation achieves the highest accuracy, but several alternative mappings perform nearly identically while reducing vocabulary size. The standard dotless mapping achieved minimal performance change (accuracy 0.938) while reducing vocabulary by 8.7\%. HWE and HCE mappings maintain virtually the same accuracy and F1-scores while reducing vocabulary by 4-6\%. LCE and LWE achieve moderate performance drops (about 1-2.7\%) with larger vocabulary reductions (up to 17\%). In summary, random character mappings achieve robust results in POS tagging, with syntactic information still effectively captured despite the remapping.

\begin{table}[htbp]
\centering
\renewcommand{\arraystretch}{1.3}

\begin{adjustbox}{max width=\linewidth}
\begin{tabular}{lccccc}
\hline
\textbf{Text Type} & \textbf{V} & \textbf{Precision} & \textbf{Recall} & \textbf{F1-score} & \textbf{Accuracy} \\
\hline
Standard Dotted   & 20,622 & 0.898 & 0.874 & 0.880 & 0.941 \\ \hline
Standard Dotless  & 18,829 & 0.898 & 0.871 & 0.881 & 0.938 \\ \hline
HCE Mapping     & 19,485 & 0.899 & 0.872 & 0.880 & 0.940 \\ \hline
LCE Mapping     & 17,891 & 0.885 & 0.866 & 0.871 & 0.933 \\ \hline
HWE Mapping       & 19,721 & 0.900 & 0.873 & 0.881 & 0.940 \\ \hline
LWE Mapping     & 17,066 & 0.876 & 0.850 & 0.856 & 0.927 \\
\hline
\end{tabular}
\end{adjustbox}
\caption{The results and vocabulary size of POS tagging for different text representation using word tokenizer.}
\label{tab:rm-pos-word-results}
\end{table}

\subsection{Named-Entity Recognition}
\label{ner-experiments}

This experiment evaluates the effect of character remapping on NER. Experiments are conducted on the ANER dataset \citep{benajiba2007anersys}, consisting of 3,972 training samples and 924 testing samples, with 10\% of the training data used for validation. The dataset includes nine entity classes, and an additional padding label is introduced during preprocessing. A BiLSTM-based model is used, with consistent hyperparameters across mappings to ensure fair comparison, as shown in Table~\ref{tab:ner_model_config} in Appendix A. Performance is evaluated primarily using token-level accuracy, along with precision, recall, and F1-score to account for class imbalance.


Table \ref{tab:rm-ner-word-results} indicates that NER performance is highly robust to character remapping. The standard dotless mapping slightly outperforms the dotted baseline while reducing vocabulary size by 7.2\%. The HWE mapping achieves the highest accuracy with a 3.2\% vocabulary reduction, suggesting that selective preservation of distinctions can improve performance. The LCE mapping maintains strong accuracy while reducing vocabulary by 13.9\%. Overall, the arbitrary character mappings do not degrade NER performance, highlighting the robustness of underlying entity representations despite vocabulary compression.

\begin{table}[htbp]
\centering
\renewcommand{\arraystretch}{1.3}

\begin{adjustbox}{max width=\linewidth}
\begin{tabular}{lccccc}
\hline
\textbf{Text Type} & \textbf{V} & \textbf{Precision} & \textbf{Recall} & \textbf{F1-score} & \textbf{Accuracy} \\
\hline
Standard Dotted   & 26,154 & 0.705 & 0.472 & 0.543 & 0.912 \\ \hline
Standard Dotless  & 24,265 & 0.702 & 0.499 & 0.563 & 0.916 \\\hline
HCE Mapping     & 24,979 & 0.694 & 0.500 & 0.566 & 0.914 \\\hline
LCE Mapping     & 22,508 & 0.719 & 0.501 & 0.575 & 0.915 \\\hline
HWE Mapping       & 25,323 & 0.713 & 0.520 & 0.584 & 0.917 \\\hline
LWE Mapping     & 22,386 & 0.694 & 0.487 & 0.559 & 0.910 \\
\hline
\end{tabular}
\end{adjustbox}
\caption{The results and vocabulary size of NER for different text representation
using word tokenizer.}
\label{tab:rm-ner-word-results}
\end{table}

\subsection{Machine Translation}
\label{mt-experiments}

We evaluate the impact of character remapping on neural machine translation (NMT). Experiments are conducted on the English-Arabic subset of the IWSLT 2017 dataset \citep{cettolo2017overview}, using 231,713 training pairs and standard validation and test splits. Text is normalized and tokenized using Moses\footnote{\url{https://github.com/hplt-project/sacremoses}}, and two tokenization strategies are explored: word-level and SentencePiece (4,000 subword units).  An encoder-decoder transformer model is trained for this task, hyperparameters are shown in Table~\ref{tab:mt_model_config} in Appendix A. BLEU score (SacreBLEU) \citep{ding2019call} is used for evaluation with greedy decoding. For the standard dotless and LCE mappings, an additional restoration step is applied after translation to reconstruct the original dotted form before evaluation, assessing reversibility and output readability.

\textbf{Arabic-to-English.} As shown in Table \ref{tab:rm-mt-bleu-vocab-combined}, SentencePiece consistently outperforms word-level tokenization across all mappings, achieving substantially higher BLEU scores. While alternative mappings achieve relatively stable BLEU scores, with only minor degradation compared to the standard dotted baseline, they reduce word-level vocabulary size by up to $\sim$19\%. HCE and HWE mappings achieve near-baseline performance while reducing vocabulary by 6-7\%, demonstrating an effective trade-off between compactness and translation quality.

\begin{table}[htbp]
\centering
\renewcommand{\arraystretch}{1.3}
\begin{adjustbox}{max width=\linewidth}
\begin{tabular}{lcccccc}
\hline
\multirow{2}{*}{\textbf{Text Type}} & 
\multicolumn{3}{c|}{\textbf{Word}} & 
\multicolumn{3}{c}{\textbf{SentencePiece}} \\
\cline{2-7}
& \textbf{V} & \textbf{Ar$\rightarrow$En} & \textbf{En$\rightarrow$Ar} &
  \textbf{V} & \textbf{Ar$\rightarrow$En} & \textbf{En$\rightarrow$Ar} \\
\hline

Standard Dotted   & 91,224 & 26.40 & 13.18 & \multirow{6}{*}{4,000} & 30.75 & 17.53 \\\cline{1-4} \cline{6-7}
Standard Dotless  & 80,866 & 25.55 & 14.03 &  & 28.95 & 17.85 \\\cline{1-4} \cline{6-7}
HCE Mapping       & 85,695 & 26.00 & 12.82 &  & 30.06 & 17.08 \\\cline{1-4} \cline{6-7}
LCE Mapping       & 73,749 & 25.35 & 13.00 &  & 28.44 & 17.54 \\\cline{1-4} \cline{6-7}
HWE Mapping       & 87,391 & 26.22 & 13.17 &  & 30.19 & 17.53 \\\cline{1-4} \cline{6-7}
LWE Mapping       & 73,788 & 25.03 & 11.81 &  & 28.13 & 17.46 \\
\hline
\end{tabular}
\end{adjustbox}
\caption{BLEU scores and Arabic vocabulary size for machine translation for different text representation using word and SentencePiece tokenizers.}
\label{tab:rm-mt-bleu-vocab-combined}
\end{table}



\textbf{English-to-Arabic.} Similar trends are observed in English to Arabic machine translation. SentencePiece again yields higher BLEU scores than word-level tokenization and shows strong robustness to vocabulary compression. Even mappings with substantial vocabulary reductions (up to $\sim$19\%) incur only modest BLEU decreases. Entropy-maximizing mappings maintain competitive performance with smaller vocabularies.




When restoration is applied to the dotless and LCE outputs, BLEU scores decrease by 1.6-3.5 points, reflecting minor reconstruction errors; detailed restoration results are shown in Table~\ref{tab:rm-mt-en2ar-bleu-restore}. However, overall translation quality remains strong, indicating that reversible remapping is feasible in practice.

\begin{table}[htbp]
\centering
\renewcommand{\arraystretch}{1.3}

\begin{adjustbox}{max width=0.8\linewidth}
\begin{tabular}{lcc}
\hline
\textbf{Text Type} & \textbf{Word} & \textbf{SentencePiece} \\
\hline
Standard Dotless  & 12.39 & 15.72 \\ \hline
LCE Mapping     & 10.43 & 14.06 \\ 
\hline
\end{tabular}
\end{adjustbox}
\caption{BLEU score of English to Arabic machine translation after restoring original representations using word and SentencePiece tokenizers.}
\label{tab:rm-mt-en2ar-bleu-restore}
\end{table}

In summary, the results indicate that machine translation performance remains robust even with random character mappings, highlighting that it is not tied to the original text representation, especially when using subword tokenization. Alternative mappings preserve competitive BLEU scores, while significantly reducing vocabulary size, and restoration enables readable final outputs with only moderate quality loss.

\section{Conclusion}
\label{sec:Conclusion}

This work investigated whether the visual structure of Arabic script reflects meaningful form-function relationships (iconicity) or whether it is largely arbitrary from an NLP perspective. 

The results indicate that Arabic NLP models do not merely tolerate historically motivated dot removal. Across the evaluated tasks, they also adapt to arbitrary but consistent reductions of the character set, indicating that downstream performance depends more on stable distributional regularities than on preserving visually meaningful letter groupings. The stability of Zipf’s Law and Heap’s Law further confirms that distributional structure remains intact despite orthographic changes.

This robustness also provides practical benefits: many mappings reduce vocabulary size and model size substantially, often with small performance losses. Vocabulary sizes were reduced by up to 50\% with minimal performance loss across multiple NLP tasks for some of the arbitrary mappings. Alternative mappings provided effective trade-offs between compression and accuracy, with HWE and HCE maintaining near-baseline performance and LCE and LWE achieving greater compression with only minor degradation. The findings suggest that, for NLP applications, preserving statistical structure is sufficient to support efficient modeling with simpler, flexible encodings.

Despite these contributions, several limitations remain. The restoration experiments employ RNN-based models, and the evaluation is limited to a selected set of NLP tasks. In addition, the extensive experimental design made large-scale evaluation with modern pre-trained large language models (LLMs) and recent neural architectures impractical given available computational resources. 

Future work may therefore investigate how contemporary pre-trained LLMs perform under alternative script mappings, explore more advanced reconstruction methods, evaluate a broader range of NLP applications, and extend the analysis to other writing systems to determine whether the observed arbitrariness generalizes across languages, potentially informing more unified and resource-efficient approaches to language representation.

\newpage

\appendix
\appendixsection{Hyperparameters}
\label{hyperparamters}

This section lists the hyperparameters of the models used in this study.

\noindent\textbf{BiLSTM-based Restoration Model.} 

\begin{table}[htbp]
\centering

\begin{tabular}{ll}
\hline
\textbf{Component} & \textbf{Value} \\
\hline
Embedding Layer Dimension & 512 \\
\hline
BiLSTM Layers & 2\\
\hline
Dropout Rate &  0.33 \\
\hline
Optimizer & Adam \\
\hline
Learning Rate & 0.001 \\
\hline
Batch Size & 256 \\
\hline
Epochs & 30 \\
\hline
Max Sequence Length & 500 characters \\
\hline
\end{tabular}
\caption{Model architecture and training configuration for restoring original text representation.}
\label{tab:model-architecture}
\end{table}

\noindent\textbf{Decoder-based Language Model.} 

\begin{table}[htbp]
\centering

\begin{tabular}{ll}
\hline
\textbf{Component} & \textbf{Value} \\

\hline
Number of Layers & 3 \\
\hline
Attention Heads per Layer & 8 \\
\hline
Feedforward Hidden Size & 2048 \\
\hline
Optimization Algorithm & RAdam \\
\hline
Epochs & 30 \\
\hline
\end{tabular}
\caption{The configurations of the transformer-based language modeling for different text representations.}
\label{tab:transformer-model}
\end{table}

\noindent\textbf{GRU-based Poem Meter Classification Model.} 

\begin{table}[H]
\centering

\begin{tabular}{ll}
\hline
\textbf{Component} & \textbf{Value} \\
\hline
Embedding size         & 256 \\ \hline
Number of layers             & 5 \\ \hline
GRU dropout            & 0.25 \\ \hline
Dense layer dropout    & 0.15 \\ \hline
Activation function    & ReLU \\ \hline
Optimizer              & Adam \\ \hline
Learning rate          & 0.001 \\ \hline
Batch size             & 512 \\ \hline
Epochs             & 50 \\ 
\hline
\end{tabular}
\caption{Model configuration for poem meter classification of different text representations.}
\label{tab:pm_model_config}
\end{table}

\noindent\textbf{GRU-based Sentiment Analysis Model.} 

\begin{table}[H]
\centering

\begin{tabular}{ll}
\hline
\textbf{Component} & \textbf{Value} \\
\hline
Embedding size       & 512 \\\hline
Number of layers & 2 \\ \hline
Number of units     & 128\\ \hline
GRU dropout          & 0.5 \\ \hline
Inter-layer dropout  & 0.5 \\ \hline
Optimizer            & Adam \\ \hline
Learning rate        & 0.01 \\\hline
Batch size           & 128  \\

\hline
\end{tabular}
\caption{Model configuration for sentiment analysis of different text representations.}

\label{tab:sa_hyperparams}
\end{table}

\noindent\textbf{BiGRU-based News Topic Classification Model.} 

\begin{table}[htbp]
\centering

\begin{tabular}{ll}
\hline
\textbf{Component} & \textbf{Value}  \\
\hline
Embedding size & 128\\ \hline
Number of layers & 2 \\ \hline
Number of units & 128\\ \hline
GRU dropout & 0.5 \\ \hline
Dense layers Dropout & 0.5 \\ \hline
Optimizer & Adam \\ \hline
Learning rate & 0.001 \\ \hline
Batch size & 128 \\
\hline
\end{tabular}
\caption{Model configuration for topic news classification of different text representations.}
\label{table:nt_model_config}
\end{table}

\noindent\textbf{BiLSTM-based POS Tagging Model.} 

\begin{table}[H]
\centering
\begin{tabular}{ll}
\hline
\textbf{Component} & \textbf{Value} \\
\hline
Embedding size        & 512 \\ \hline
Recurrent layers      & 5  \\ \hline
LSTM hidden units     & 512 \\ \hline
Dropout rate          & 0.5 \\ \hline
Optimizer             & Adam \\ \hline
Learning rate         & 0.001 \\ \hline
Batch size            & 256 \\ \hline
Sequence length       & 400 tokens \\
\hline
\end{tabular}
\caption{Model configuration for POS tagging of different text representations.}
\label{tab:pos_model_config}
\end{table}

\noindent\textbf{BiLSTM-based NER Model.} 

\begin{table}[H]
\centering
\begin{tabular}{ll}
\hline
\textbf{Component } & \textbf{Value} \\
\hline
Embedding size         & 256 \\ \hline
Recurrent layers       & 3  \\ \hline
LSTM hidden units      & 512 \\ \hline
Dropout rate    & 0.5 \\ \hline
Optimizer              & Adam \\ \hline
Learning rate          & 0.001 \\ \hline
Batch size             & 256 \\ \hline
Sequence length        & 220 tokens \\
\hline
\end{tabular}
\caption{Model configuration for NER of different text representations.}
\label{tab:ner_model_config}
\end{table}

\noindent\textbf{Encoder-decoder Machine Translation Model.} 

\begin{table}[htbp]
\centering

\begin{tabular}{ll}
\hline
\textbf{Component} & \textbf{Value} \\
\hline
Embedding size             & 512 \\ \hline
Encoder / Decoder layers   & 6 / 6 \\ \hline
Feedforward hidden size    & 2048 \\ \hline
Attention heads            & 8 \\ \hline
Optimizer                  & RAdam \\ \hline
Learning rate              & 0.0001 \\
\hline
\end{tabular}
\caption{Model configuration for machine translation of different text representations.}
\label{tab:mt_model_config}
\end{table}

\clearpage

\appendixsection{Additional Tokenizers Analysis}
\label{analysis2}


\noindent\textbf{Farasa Morphological Tokenizer.} 

\begin{table}[ht]
\centering
\begin{adjustbox}{max width=\linewidth}
\begin{tabular}{lrrrrc}
\hline
\multicolumn{1}{c}{\textbf{Representation}} & 
\multicolumn{1}{c}{\textbf{N}} & 
\multicolumn{1}{c}{\textbf{V}} & 
\multicolumn{1}{c}{\textbf{V/N}} & 
\multicolumn{1}{c}{\textbf{Avg. N}} & 
\multicolumn{1}{c}{\textbf{Entropy}} \\
\hline
Standard Dotted   &\multirow{6}{*}{280,636,479} & 1,134,585 & 0.0040  & 7.3107 & 9.4150  \\ \cline{1-1} \cline{3-6}
Standard Dotless  &&   891,387 & 0.0032 & 7.7650  & 9.2203 \\ \cline{1-1} \cline{3-6}
HCE Mapping     && 1,001,373 & 0.0036 & 7.5874 & 9.2905 \\ \cline{1-1} \cline{3-6}
LCE Mapping     &&   798,443 & 0.0028 & 7.9618 & 8.8594 \\ \cline{1-1} \cline{3-6}
HWE Mapping       && 1,017,451 & 0.0036 & 7.5630  & 9.3247 \\ \cline{1-1} \cline{3-6}
LWE Mapping     &&   728,374 & 0.0026 & 8.1102 & 8.8953 \\
\hline
\end{tabular}
\end{adjustbox}
\caption{Farasa morphological tokenizer statistics for Wikipedia Arabic corpus using different text representations.}
\label{tab:rm-analysis-farasa}
\end{table}

\noindent\textbf{Disjoint-Letters Tokenizer.} 

\begin{table}[ht]
\centering

\begin{adjustbox}{max width=\linewidth}
\begin{tabular}{lccccr}
\hline
\multicolumn{1}{c}{\textbf{Representation}} & 
\multicolumn{1}{c}{\textbf{N}} & 
\multicolumn{1}{c}{\textbf{V}} & 
\multicolumn{1}{c}{\textbf{V/N}} & 
\multicolumn{1}{c}{\textbf{Avg. N}} & 
\multicolumn{1}{c}{\textbf{Entropy}} \\
\hline
Standard Dotted    &\multirow{2}{*}{416,084,771} & 248,567 & 0.0006 & 5.5916 & 8.0575 \\ \cline{1-1} \cline{3-6}
Standard Dotless  && 112,688 & 0.0003 & 6.1118 & 7.4495 \\ \hline
HCE Mapping     &348,449,648 & 351,116 & 0.0010  & 6.8020  & 8.5234 \\ \hline
LCE Mapping     &337,284,590 & 163,735 & 0.0005 & 7.0911 & 8.0597 \\ \hline
HWE Mapping       &436,782,517 & 205,954 & 0.0005 & 6.1467 & 7.3322 \\ \hline
LWE Mapping     &252,285,858 & 450,082 & 0.0018 & 7.9454 & 10.0654 \\
\hline
\end{tabular}
\end{adjustbox}
\caption{Disjoint-letters tokenizer statistics for Wikipedia Arabic corpus using different text representations.}
\label{tab:rm-analysis-disjoint}
\end{table}

\noindent\textbf{Zipf's Law.} 

\begin{figure}[H]
    \centering
    \includegraphics[width=0.75\linewidth]{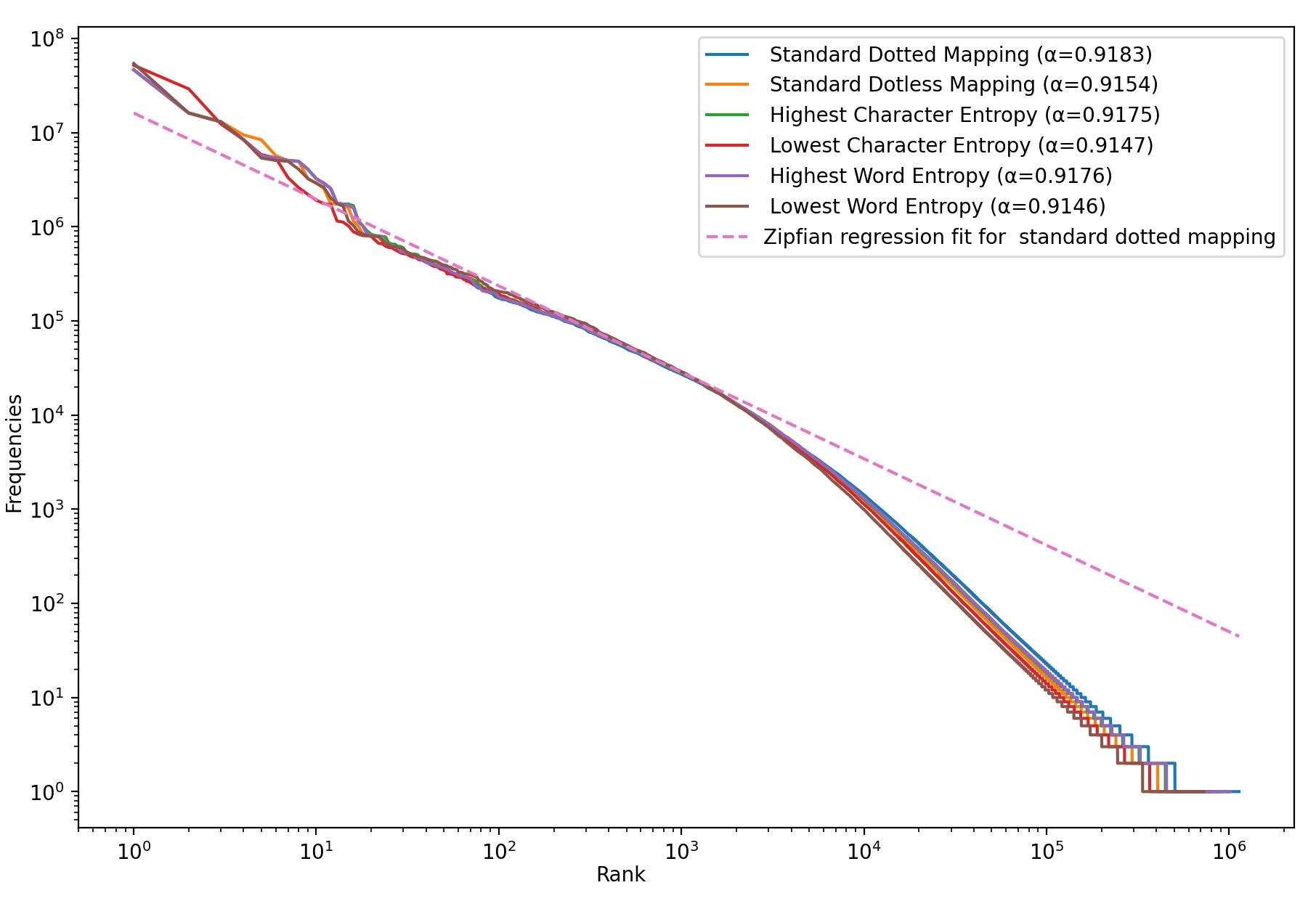}
    \caption{Zipf’s Law plot under Farasa tokenizer for various character mappings.}
    \label{fig:rm-zipfs-farasa}
\end{figure}

\begin{figure}[H]
    \centering
    \includegraphics[width=0.75\linewidth]{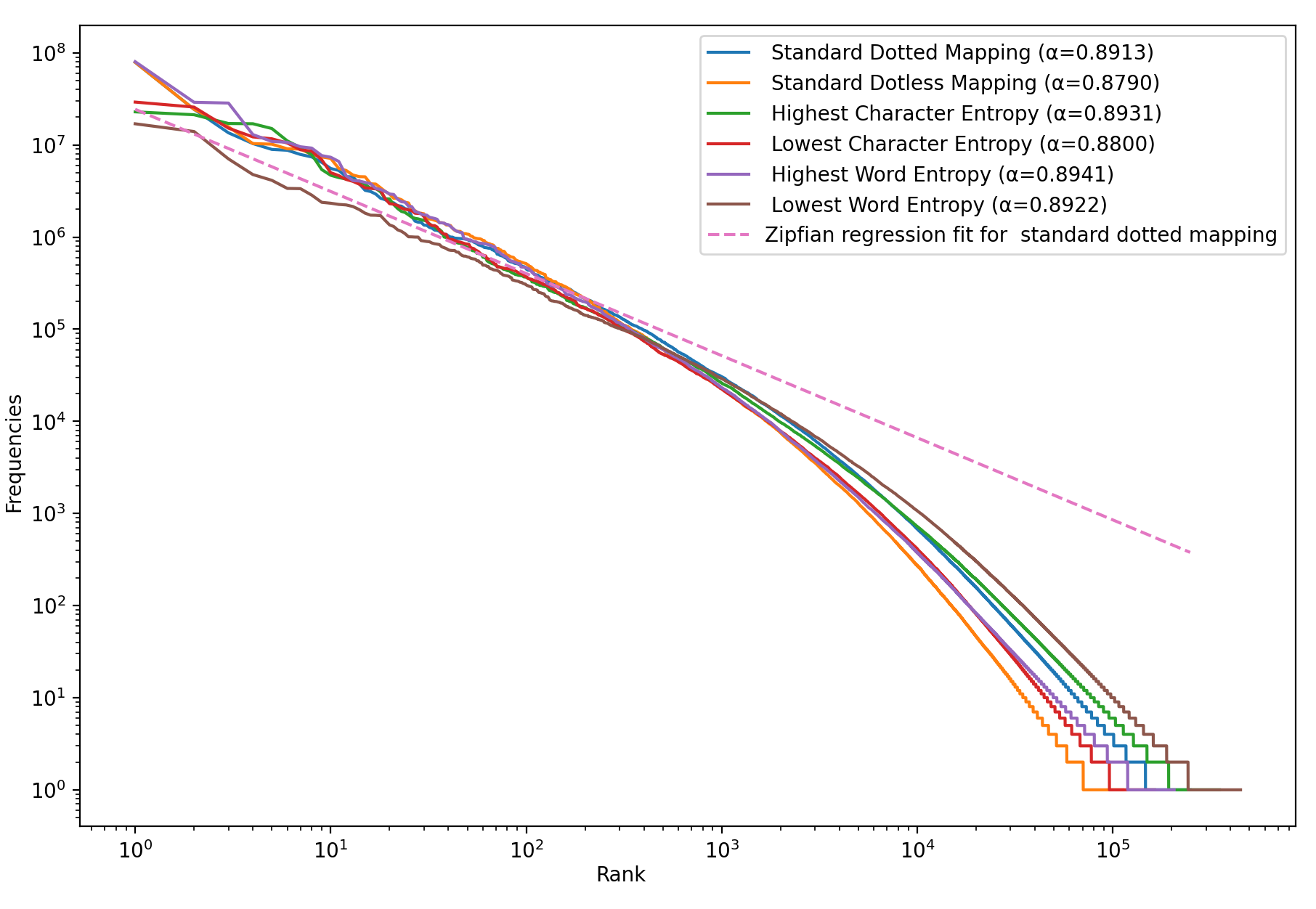}
    \caption{Zipf’s Law plot under disjoint tokenizer for various character mappings.}
    \label{fig:rm-zipfs-disjoint}
\end{figure}

\noindent\textbf{Heap's Law.} 

\begin{figure}[H]
    \centering
    \includegraphics[width=0.75\linewidth]{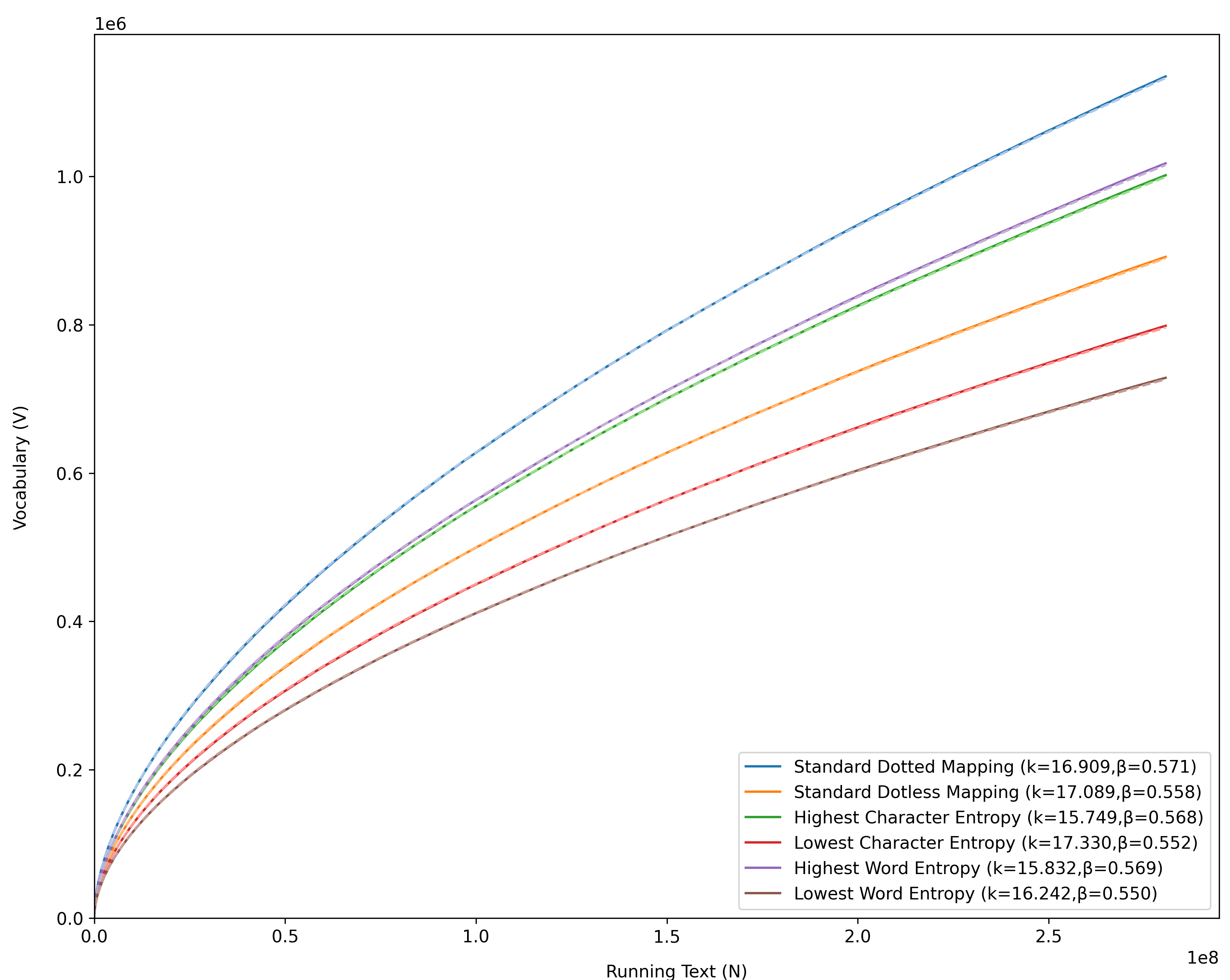}
    \caption{Heap’s Law plot under Farasa tokenizer for various character mappings, (dashed lines represent the regression fit for each representation).}
    \label{fig:rm-heaps-farasa}
\end{figure}

\begin{figure}[ht]
    \centering
    \includegraphics[width=0.75\linewidth]{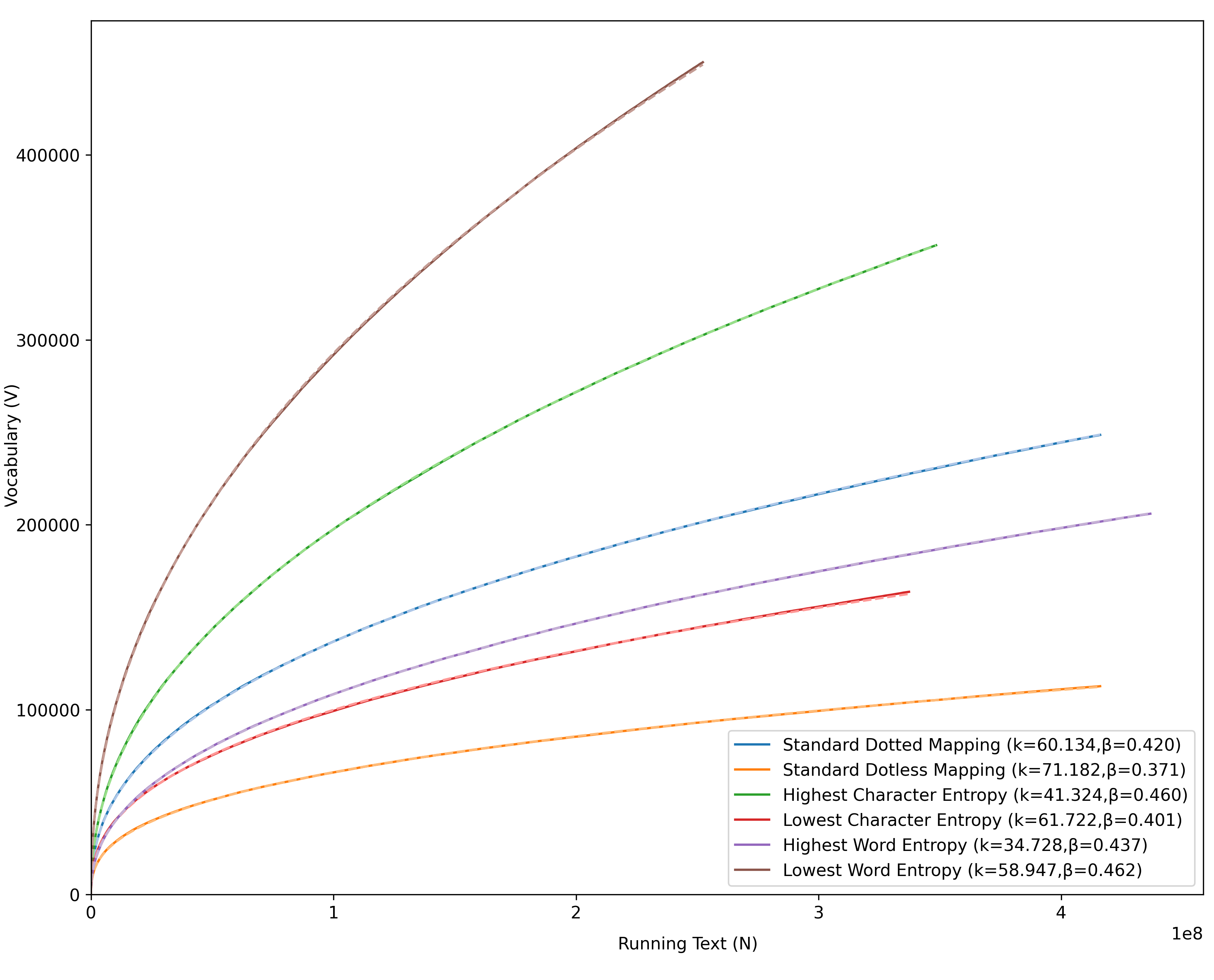}
    \caption{Heap’s Law plot under disjoint tokenizer for various character mappings, (dashed lines represent the regression fit for each representation).}
    \label{fig:rm-heaps-disjoint}
\end{figure}

\newpage
\starttwocolumn

\begin{acknowledgments}
The authors would like to thank King Fahd University of Petroleum and Minerals (KFUPM) for supporting this work. Irfan Ahmad would like to additionally thank Saudi Data and AI Authority (SDAIA) and KFUPM for supporting him through SDAIA-KFUPM Joint Research Center for Artificial Intelligence grant number JRC--AI--RFP--10.
\end{acknowledgments}


\bibliographystyle{compling}
\bibliography{COLI_template}

\end{document}